\documentclass[journal]{IEEEtran}
\usepackage{amsmath,amsfonts}

\usepackage{algorithm}
\usepackage{array}
\usepackage{textcomp}
\usepackage{stfloats}
\usepackage{url}
\usepackage{verbatim}
\usepackage{graphicx}
\usepackage{cite}
\usepackage{amsmath}
\usepackage{amssymb}
\usepackage{booktabs}

\usepackage{multirow}

\usepackage{algorithmicx}
\usepackage{algpseudocode}
\floatname{algorithm}{Algorithm}

\usepackage{xcolor,colortbl}
\usepackage{subfig}
\definecolor{darkred}{rgb}{0.70, 0.0, 0.0}

\newcommand{\best}[1]{{{\textbf{#1}}}}
\newcommand{\second}[1]{{\textcolor{blue}{{#1}}}}

\usepackage{balance}

\definecolor{lightblue}{RGB}{57,89,163}

\begin{document}

\title{EPS: Efficient Patch Sampling for Video Overfitting \\ in Deep Super-Resolution Model Training}

\author{
Yiying Wei,~\IEEEmembership{Student Member,~IEEE},
Hadi Amirpour,~\IEEEmembership{Member,~IEEE},
Jong Hwan Ko,~\IEEEmembership{Member,~IEEE},
and Christian Timmerer,~\IEEEmembership{Senior Member,~IEEE}
\thanks{This work was supported in part by the Austrian Federal Ministry for Digital and Economic Affairs; in part by the National Foundation for Research, Technology, and
Development; in part by the Christian Doppler Research Association; and in
part by the Christian Doppler Laboratory ATHENA.}% 
\thanks{Yiying Wei, Hadi Amirpour, and Christian Timmerer are with the Christian Doppler Laboratory ATHENA, Alpen-Adria-Universität Klagenfurt, 9020 Klagenfurt,
Austria (e-mail: yiying.wei@aau.at; hadi.amirpour@aau.at; christian.timmerer@aau.at).}
\thanks{Jong Hwan Ko is with the College of Information and Communication Engineering, Sungkyunkwan University, 16419 Suwon, South Korea (e-mail: jhko@skku.edu).}
% \thanks{Manuscript received April 19, 2021; revised August 16, 2021.}
}

% The paper headers
\markboth{Journal of \LaTeX\ Class Files,~Vol.~14, No.~8, May~2025}%
{Yiying Wei, Hadi Amirpour, \MakeLowercase{\textit{et al.}}: EPS: Efficient Patch Sampling for Video Overfitting in Deep Super-Resolution Model Training}

\maketitle

\IEEEpubid{\begin{tabular}[t]{@{}c@{}}Copyright \copyright~2026 IEEE. Personal use of this material is permitted. However, permission to use this material \\ for any other purposes must be obtained from the IEEE by sending an email to pubs-permissions@ieee.org.\end{tabular}}

\begin{abstract}
Leveraging the overfitting property of deep neural networks (DNNs) is trending in video delivery systems to enhance video quality within bandwidth limits. Existing approaches transmit overfitted super-resolution (SR) model streams for low-resolution (LR) bitstreams, which are used to reconstruct high-resolution (HR) videos at the decoder. Although these approaches show promising results, the huge computational costs of training a large number of video frames limit their practical applications. To overcome this challenge, we propose an efficient patch sampling method named \textbf{EPS} for video SR network overfitting, which identifies the most valuable training patches from video frames. 
To this end, we first present two low-complexity Discrete Cosine Transform (DCT)-based spatial-temporal features to measure the complexity score of each patch directly. By analyzing the histogram distribution of these features, we then categorize all possible patches into different clusters and select training patches from the cluster with the highest spatial-temporal information. The number of sampled patches is adaptive based on the video content, addressing the trade-off between training complexity and efficiency. 
Our method reduces the number of training patches by 75.00\% to 91.69\%, depending on the resolution and number of clusters, while preserving high video quality and greatly improving training efficiency. Our method speeds up patch sampling by up to 82.1$\times$ compared to the state-of-the-art patch sampling technique (EMT). %The source code will be made available upon acceptance of the paper.
\end{abstract}

\begin{IEEEkeywords}
Patch sampling, data overfitting, super resolution, model training.
\end{IEEEkeywords}

\section{Introduction}
\label{sec:intro}

\IEEEpubidadjcol

\IEEEPARstart{W}ith the ever-increasing amount of video content, video applications, and the ongoing evolution of video in various dimensions, such as spatial resolution and temporal resolution (frame rate), transmitting high-quality, high-resolution videos presents a significant challenge. In response to these challenges, new video codecs have been introduced, such as Versatile Video Coding (VVC)~\cite{VVC_overview_2021} or AOMedia Video 1 (AV1)~\cite{AV1_overview_2020}, which employ more efficient compression techniques to help transmit high-quality video content while reducing bandwidth requirements. However, these video coding methods still face limitations to further improve compression performance, as they rely on hand-crafted techniques and highly engineered modules. 

\IEEEpubidadjcol

With the development of deep learning, leveraging deep neural networks (DNNs) to enhance video compression has become a new trend in modern video transmission systems. Numerous learning-based video compression methods~\cite{DeepCoder_chen_2017, ma2019image_neural_review, shen_codingsr_2011, li2018_cnn_coding,VCII_wu_2018, content_lu_2020, UAR-NVC_2026, MARINA_2026} have been proposed to deliver high-quality video streams to users. Among these approaches, an emerging number of approaches integrate super-resolution (SR) techniques to reduce bandwidth requirements~\cite{fischer_flownet_2015,lin2019_cnn_sr_hevc,afonso2019_compression_resolution, habibian_video_2019,rippel_learned_2019}. These methods transmit low-bitrate low-resolution (LR) videos and super-resolve them to high-resolution (HR) videos on the end-user device by applying pre-trained SR models. These SR models are typically trained on a limited dataset and may encounter difficulties adapting to new video content. However, creating a universal DNN model that excels with all Internet videos is impractical. To overcome this limitation, recent advances in neural-enhanced video delivery~\cite{NAS_2018, LiveNAS_2020,SRVC_2021,CaFM_liu_2021, DeepStream_2025} leverage the overfitting property of DNNs to achieve quality improvements. These approaches train an SR model for \textbf{each video} and stream the LR video along with the corresponding content-aware SR model to the end-user device. The reinforced expressive power of content-aware SR models significantly improves the quality of resolution-upscaled videos. 

\begin{figure*}[!t]
\centering
\includegraphics[width=0.85\textwidth]{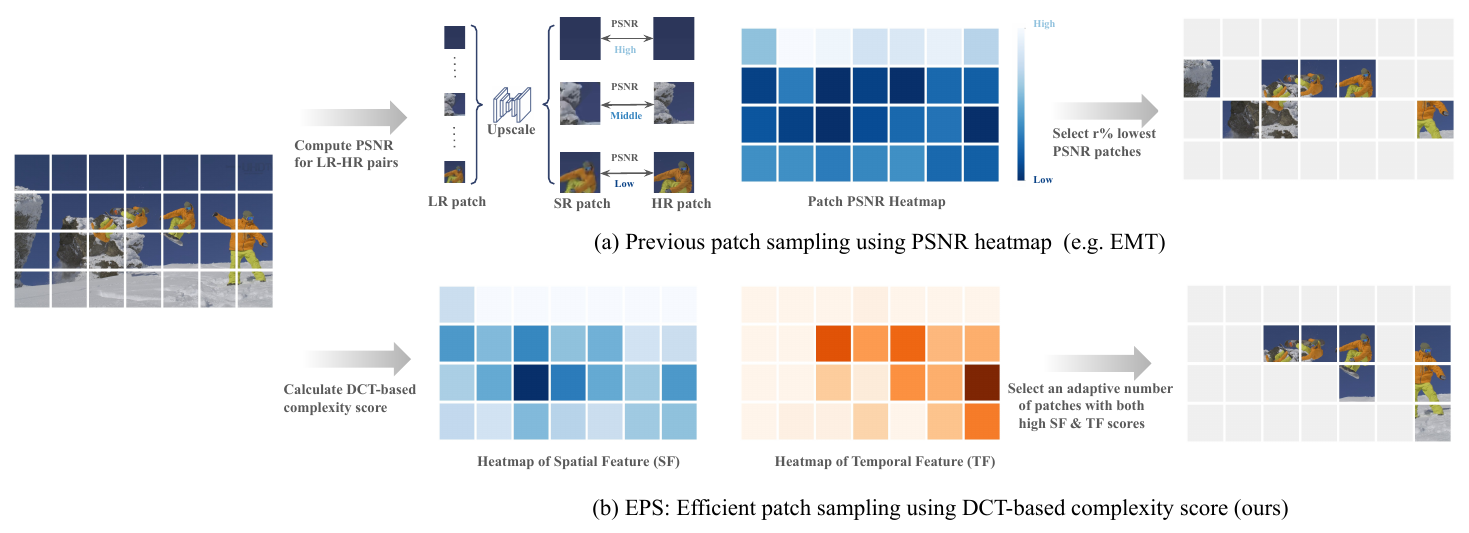}
\caption{Comparison of different patch sampling methods. (a) State-of-the-art patch sampling methods compute PSNR for each LR-HR patch pair to extract the informative patches. (b) \textbf{EPS}: Efficient patch sampling directly uses DCT-based complexity scores (\textit{SF} and \textit{TF}) to select informative patches. The \textit{SF} score indicates the complexity of the texture information within the patch itself, while the \textit{TF} score indicates movement and change between frames to help reduce temporal redundancies for patch sampling. 
}
\label{fig:conceptual_figure}
\end{figure*}

Recent advancements in the broader super-resolution (SR) domain for both image~\cite{bergner2022ips, wang2024camixersr, yi2025rsisr} and video~\cite{yi2021omniscient, aesr3d, yi2024, yi2025, feng2023deep, temp_2023} have introduced increasingly sophisticated architectures that significantly improve reconstruction fidelity. These methods mainly focus on advancing network architectures and modeling capabilities to better restore high-frequency details. However, such improvements typically require substantial computational resources and extensive training procedures. Consequently, although neural-enhanced video delivery shows promising performance, the computational cost of training content-aware SR models remains a major barrier to practical deployment. In many neural video delivery frameworks, SR models must be adapted or fine-tuned for individual video streams, which makes training efficiency a critical concern. As video resolutions continue to increase and large-scale video processing pipelines become more common, the associated training workload grows significantly and can hinder real-time or large-scale deployment. Additionally, deploying such models for large-scale video processing and delivery workflows may lead to considerable energy consumption, which raises concerns regarding sustainability and environmental impact~\cite{afzal_survey_2024}.

To reduce the computational cost of network training, EMT~\cite{EMT_2022} proposed a patch sampling method to select the most informative patches using a \textit{patch PSNR heatmap}, showing training gains comparable to using all frames. Specifically, it uses (\textit{i}) a pre-trained SR model to super-resolve all LR patches of one frame, and then (\textit{ii}) calculates their PSNR values with the original HR patches to generate the patch PSNR heatmap. As shown in Fig.~\ref{fig:conceptual_figure}(a), the patch PSNR heatmap varies across different patch locations according to its content. Regions with complex textures usually represent lower PSNR values than smooth regions because they are more challenging to restore with an SR model. 
The patch PSNR heatmap indeed partially reflects the texture complexity of patches, assisting in the identification of valuable patches for training content-aware models. However, the current state-of-the-art patch sampling methods, such as EMT~\cite{EMT_2022}, still have two main drawbacks: 
\begin{itemize}
    \item First, generating patch PSNR heatmaps for all frames is time-consuming. It requires additional computational resources, as it involves the inference of a DNN and calculating PSNR for each patch, which online training attempts to avoid.
    \item Second, these methods sample patches only based on the SR quality comparisons without considering the temporal redundancy between frames. When temporal complexity is low -- indicating that a patch is similar to its co-located patch in the previous frame -- it can be excluded from the training set due to the redundancy, thus reducing unnecessary computational load.
\end{itemize}

In this paper, we propose Efficient Patch Sampling (EPS) for high-quality and efficient video super-resolution. As shown in Fig.~\ref{fig:conceptual_figure}(b), our method leverages spatial-temporal information to quickly select the most informative patches from video frames without the need to super-resolve frames and calculate the quality. We introduce two DCT-based features to directly evaluate the spatial and temporal complexity of patches in LR video frames. Compared to PSNR heatmaps that rely on DNN inference and are calculated on patches after SR by comparing them to the corresponding HR patches, DCT is a low-complexity computation on LR patches that enables faster execution on both CPU and GPU, significantly speeding up informative patch scoring. Instead of simply selecting the complex patches of each frame, we sample the patches by considering both temporal and spatial dimensions as the training set for the content-aware SR model. Our approach excludes relatively static patches across frames from repeated training, thereby reducing temporal redundancies. 

Our contributions can be summarized as follows:
\begin{itemize}
\setlength\leftskip{-1em}
    \item We introduce two low-complexity DCT-based informative features to measure the spatial-temporal complexity of each patch. This approach is fast and effective in guiding the selection of the most informative patches, maximizing the content-aware training gain as rapidly as possible. Compared to EMT, which relies on heavy DNN inference to generate PSNR heatmaps, our approach accelerates the patch selection process by up to 82.1$\times$ (\textit{cf.} Section~\ref{sec:patch-feature}).
    
    \item We present a novel patch sampling algorithm for content-aware video SR training, which utilizes the histogram distribution of patch features for clustering to select patches with the highest spatial-temporal information. The number of sampled patches is content-adaptive, effectively addressing the trade-off between training cost and efficiency. This allows our method to dynamically reduce the training data by 75.00\% to 91.69\% while preserving high SR video quality (\textit{cf.} Section~\ref{sec:patch-sampling}).
    
    \item While prior methods evaluate patch importance strictly on a per-frame spatial basis, our EPS method explicitly incorporates temporal dynamics. By evaluating temporal information, our method effectively removes redundant patches across consecutive frames, further enhancing training efficiency (\textit{cf.} Section~\ref{sec:eps}).
    
    \item We conduct comprehensive experiments based on various SR architectures to evaluate the advantages and generalization of our method (\textit{cf.} Section~\ref{sec:experiments}).
\end{itemize}

\section{Related Work}
\subsection{Content-aware Neural Video Delivery}
% \noindent\textbf{Content-aware Neural Video Delivery.}
NAS~\cite{NAS_2018} was one of the first neural-enhanced video delivery frameworks proposed to integrate a per-video SR model. A DNN is trained for each LR video content, and both the LR video and its associated DNN are delivered to the client side, which are jointly used to enhance its quality. LiveNAS~\cite{LiveNAS_2020} proposed a live video ingest framework that integrates an online training module into the original NAS approach~\cite{NAS_2018}. However, content-aware SR models with large parameters still introduce an overhead to the delivery process. %CaFM~\cite{CaFM_liu_2021} divides a lengthy video into several chunks and employs a joint learning technique to reduce the necessary streams of model parameters. 
SRVC~\cite{SRVC_2021} encodes a video into content streams and time-varying model streams, updating only a fraction of the model parameters over video chunks to better handle the available bandwidth budget. DeepStream~\cite{DeepStream_2025} utilizes compressed content-aware SR networks to achieve significant bitrate savings while maintaining the same quality for end-user devices with GPU capabilities. Nevertheless, these approaches still demand significant computational resources for training the network; however, utilizing patch sampling can mitigate this requirement.

\subsection{Patch Sampling}
% \noindent\textbf{Patch Sampling.}
In LiveNAS's online training module~\cite{LiveNAS_2020}, fixed-size LR-HR patches are sampled from \textbf{random} locations in a given frame. However, employing a random sampling method may lead to selecting suboptimal patches for training. In practice, more complex patches in texture may contribute significantly to model training~\cite{empirical_toneva_2018, LiveNAS_2020, wang2021samplingaug, mest_yuan_2021}. EMT~\cite{EMT_2022} used the PSNR of the HR patch and the decoded LR patch upscaled by the SR model, demonstrating higher precision in identifying complex patches that yield greater training gains. However, evaluating an SR model on all possible patch pairs is time-consuming and brings additional costs.

Furthermore, existing methods often fail to effectively exploit spatial information within individual frames and temporal information across consecutive frames. Ignoring spatial-temporal information can result in suboptimal and redundant training data, thereby reducing the training efficiency. Our work differs from these works by sampling complex patches using simple, yet efficient, DCT-based features that account for both spatial and temporal information. We note that some efficient SR methods~\cite{kong2021classsr, STDO_2023} also leverage the patch complexity. However, these methods aim to accelerate execution speed by training all patches with SR models at multiple scales to handle various patch complexities. Therefore, we do not compare our patch sampling method with these works, as data sampling can further accelerate them.

\begin{figure*}[h]
  \centering
  \includegraphics[width=0.82\linewidth]{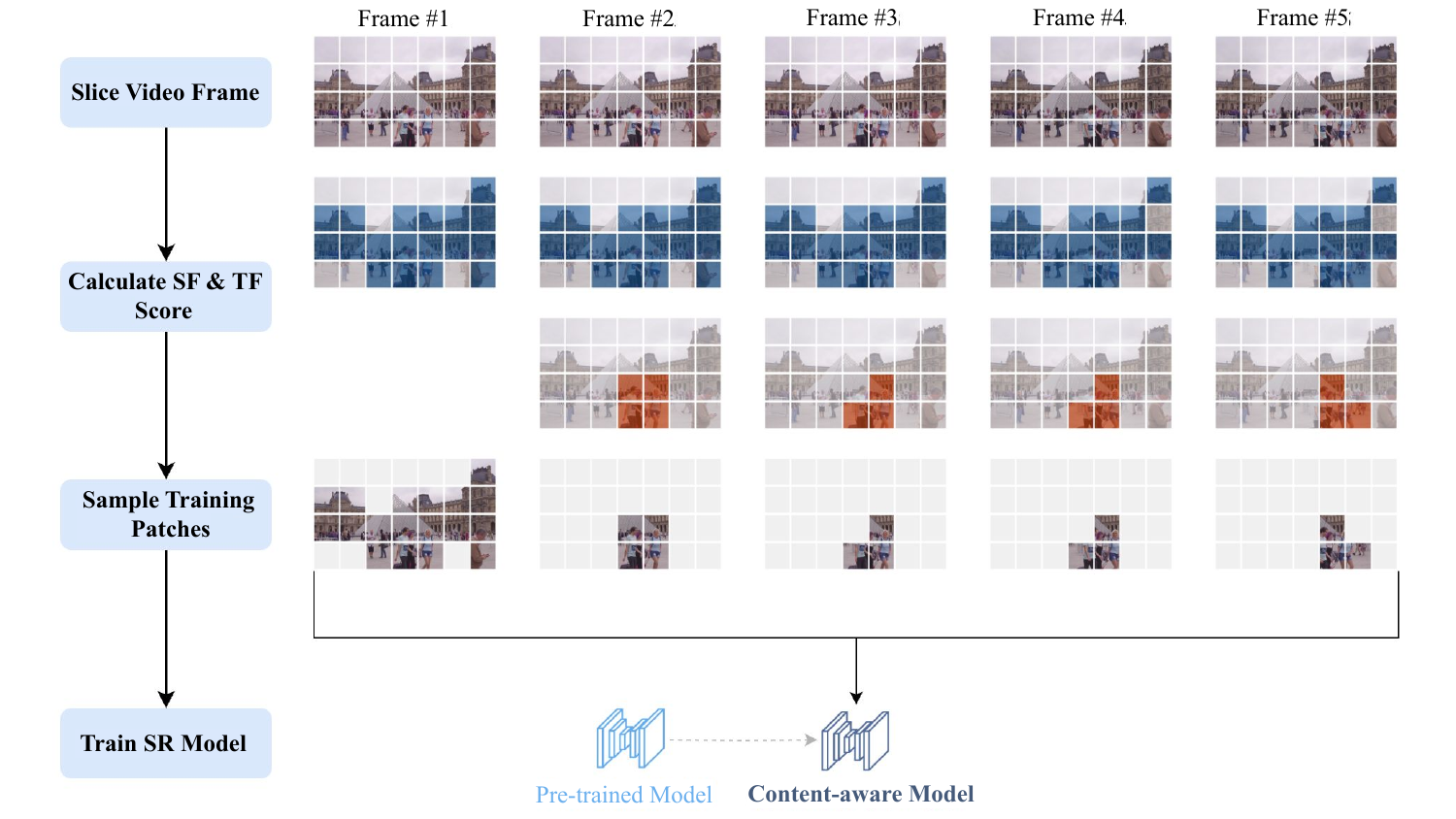}
  \caption{Overview of EPS method. Each video frame is sliced into patches. The informative complexity of each patch is determined by spatial features (\emph{SF}) and temporal features (\emph{TF}). For each frame, we group all patches into $N$ clusters based on the histogram distribution of feature scores and select the cluster of \textbf{highest} spatial-temporal information for training a content-aware SR model (using a pre-trained model as a basis). In the figure, we set the number of clusters to two for better readability. The blue and orange patches represent clusters with high \textit{SF} and \textit{TF} scores, respectively.
  }
  \label{fig:overview}
\end{figure*}

\section{Efficient Patch Sampling (EPS)}
\label{sec:eps}

In this section, we present our efficient patch sampling method to accelerate the training with reduced computational costs. Fig.~\ref{fig:overview} shows an overview of our proposed method. We first split the LR video into frames and divide each frame into a grid of non-overlapping patches. Our method leverages both spatial and temporal features to evaluate the texture complexity of each patch. We then group all patches into $N$ clusters according to the histogram distribution of our proposed features. The patches of the cluster with the \textbf{highest} spatial-temporal information are selected to train a content-aware SR model. We first introduce the features to evaluate the patch complexity in both spatial and temporal dimensions in Section~\ref{sec:patch-feature}. We then propose a patch sampling algorithm to determine the most valuable training patches for the SR model training in Section~\ref{sec:patch-sampling}.

\subsection{Patch Features}
\label{sec:patch-feature}

Related work~\cite{LiveNAS_2020, wang2021samplingaug, EMT_2022} shows that more informative patches provide higher training gains than others. Given that not all parts of the video are equally important for training, patch sampling aims to quickly select challenging patches and discard uninformative or redundant patches. 
To efficiently sample patches that achieve the goal, we introduce two informative features for each patch: (\textit{i}) Spatial Feature (\textit{SF}) and (\textit{ii}) Temporal Feature (\textit{TF}).

From a signal processing perspective, when an image or video is \textit{downsampled} and \textit{compressed}, high-frequency information, such as edges, fine textures, and rapid spatial variations, is attenuated or removed. This occurs because practical systems apply low-pass filtering prior to decimation to prevent aliasing, and compression schemes quantize high-frequency components more aggressively to reduce bitrate. Formally, downsampling by a factor $M$ compresses and aliases the spectrum as
\begin{equation}
Y(e^{j\omega}) = \frac{1}{M}\sum_{k=0}^{M-1} 
X\!\left(e^{j\frac{\omega + 2\pi k}{M}}\right),
\end{equation}
which shows that frequency components beyond the reduced Nyquist limit overlap and cannot be uniquely recovered. Consequently, reducing spatial resolution inherently restricts the maximum representable frequency and discards high-frequency content exceeding this limit. As a result, simple interpolation methods, such as bicubic or Lanczos, are sufficient for smooth, low-frequency patches that remain approximately bandlimited within the reduced Nyquist range. In contrast, complex patches containing rich edges and fine details lose substantial high-frequency information during downsampling and compression, and recovering these details requires learned priors and nonlinear inference, which is the fundamental role of super-resolution networks that synthesize plausible high-frequency content rather than merely interpolate missing samples.

\begin{figure}[t]
  \centering
  \includegraphics[width=0.48\textwidth]{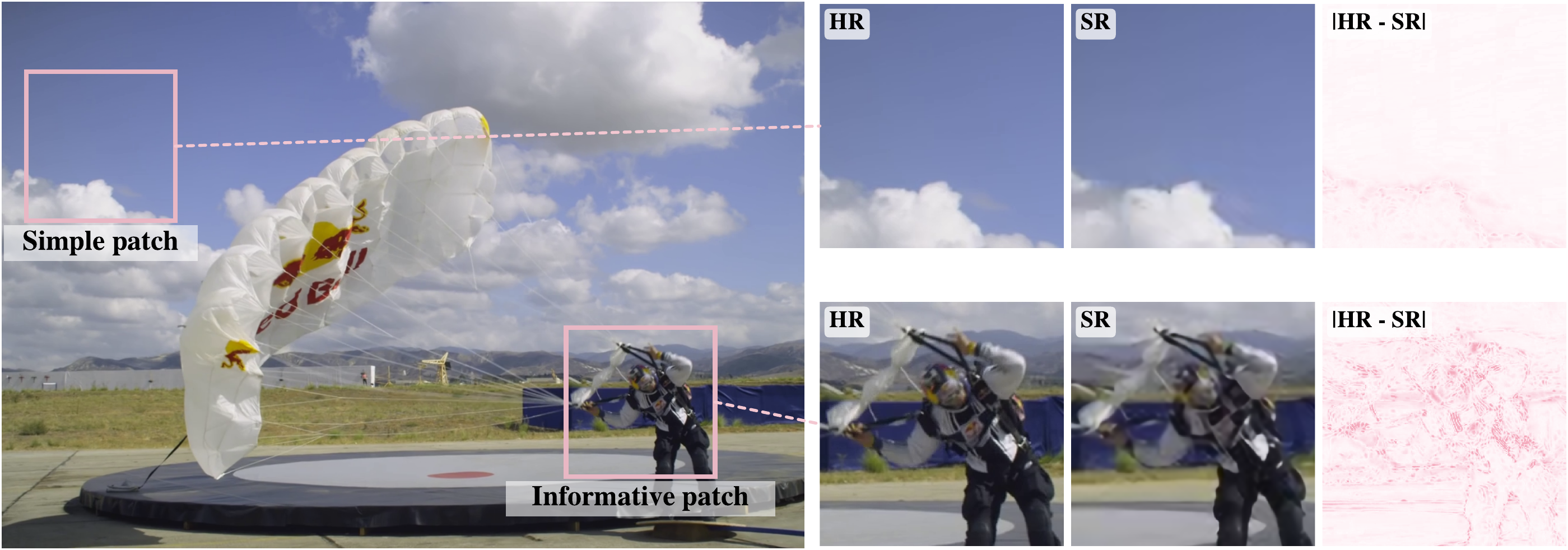}
  \caption{Comparison of simple and informative patches following super-resolution. Given an HR patch, we apply WDSR~\cite{WDSR_2018} to super-resolve its $4\times$ LR counterpart into an SR patch, and visualize the absolute pixel-wise difference between the HR and SR patches.}
  \label{fig:compare_simple_complex_patch}
\end{figure}

As shown in Fig.~\ref{fig:compare_simple_complex_patch}, high-frequency information is clearly lost in the patch with fine textures. Therefore, patch complexity and informativeness are intrinsically linked to their frequency content. From an information-theoretic perspective, patches with higher high-frequency energy contain greater structural entropy and larger reconstruction uncertainty after downsampling, making them more informative for learning. Therefore, we introduce the \textit{Spatial Feature (SF)} as a frequency-aware measure of patch complexity. Following prior DCT-based texture analysis works~\cite{E_DCT_2007,menon_vca_2022}, $SF$ is computed in the DCT domain using an exponentially weighted energy formulation that assigns greater emphasis to higher-frequency coefficients:

\begin{equation}
    SF = \sum_{i=0}^{w-1}\sum_{j=0}^{h-1}e^{[(\frac{ij}{wh})^2-1]}|{DCT(i, j)}|
    \label{eq:SF}
\end{equation}

where \textit{w} and \textit{h} are the width and height of the patch, and $DCT(i, j)$ is the \textit{$(i, j)^{th}$} DCT component when \textit{$i + j > 0$}, and 0 otherwise. This exponential weighting reflects the fact that high-frequency coefficients are more severely degraded by downsampling and compression, and therefore contribute more to reconstruction difficulty and training gain. Patches with higher $SF$ values correspond to regions where interpolation is insufficient and nonlinear modeling is necessary, making them inherently more informative for super-resolution training.

Furthermore, consecutive video frames often contain co-located patches with highly similar frequency representations. We incorporate the \textit{Temporal Feature (TF)} to quantify the temporal variation between video frames.  Formally, we denote the total $T$ frames of the given LR video as $I_{1}$, $I_{2}$, ..., $I_{T}$. For a patch of frame $I_t$ ($1 < t \leq T$), the $TF$ defines the complexity of the temporal variation between video frames and is computed as the difference of the DCT component of each patch of the current frame compared to its previous frame:

\begin{small}
\begin{equation}
    TF_{t} =
    \sum_{i=0}^{w-1}\sum_{j=0}^{h-1}e^{[(\frac{ij}{wh})^2-1]}|{DCT(i, j)}_t - {DCT(i, j)}_{t-1}| 
    \label{eq:TF}
\end{equation}
\end{small}

A low $TF$ value indicates that the two patches provide nearly identical supervisory signals. By filtering out patches with low $TF$ scores, we reduce redundant gradients and ensure the training focuses on temporally informative regions exhibiting meaningful structural changes. Together, $SF$ and $TF$ provide a theoretically grounded mechanism for selecting patches with high reconstruction uncertainty and low redundancy, thereby identifying the most informative samples for super-resolution learning.

\begin{figure}[tb]
  \centering
  \includegraphics[width=0.96\linewidth]{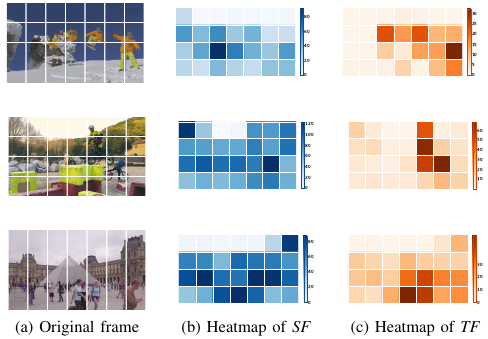}
  \caption{Example heatmaps of \textit{SF} and \textit{TF} scores of the video frames from the Inter4K dataset~\cite{Inter4K_dataset}.}
  \label{fig:heatmap_sf_tf}
\end{figure}

Example heatmaps of $SF$ and $TF$ are shown in Fig.~\ref{fig:heatmap_sf_tf} ($w = 64$, $h=64$). A high $SF$ score represents complex texture and rich patch information. Consequently, a high $TF$ score indicates that the patch in frame $I_t$ has obvious changes compared to $I_{t-1}$. Therefore, the $TF$ serves as an indicator of redundancy in co-located patches across frames.

\begin{figure}[tb]
  \centering
   \includegraphics[width=0.96\linewidth]{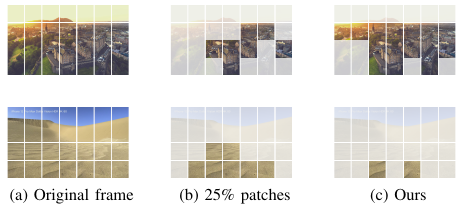}
  \caption{Example of two different patch sampling algorithms. Selecting the top $r\%$ of patches may either miss important information or lead to redundant training on similar patches, while our sampling algorithm mitigates these issues.}
  \label{fig:example_sampling}
\end{figure}

\subsection{Patch Sampling Algorithm}
\label{sec:patch-sampling}
In this section, we propose a spatial-temporal patch sampling algorithm. Previous works~\cite{LiveNAS_2020, EMT_2022} have relied on setting fixed thresholds for sampling simply by selecting the top $r\%$ of patches according to their information complexity. However, such sampling methods do not take into account the varied distribution of patch information complexity across different video contents. For individual videos, this direct sampling approach may overlook important information or overly train on simpler information. As shown in Fig.~\ref{fig:example_sampling}, in a complex video, sampling 25\% of patches might still exclude informative patches, whereas, in a simple video, the same percentage could include non-informative content. Therefore, we utilize the histogram distribution of spatial-temporal features for clustering to conduct patch sampling.
Algorithm~\ref{alg:patch_sampling_strategy} shows our patch sampling algorithm, which selects patches with the highest spatial-temporal information and discards uninformative or redundant patches. The goal is to sample LR-HR patch pairs $P$ to train a content-aware SR model of a $T$-frame video sequence.

\renewcommand{\thealgorithm}{1}
\begin{algorithm}[tb]
\footnotesize
\caption{Patch Sampling Strategy}
\label{alg:patch_sampling_strategy}
\begin{algorithmic}[1]
\Require Frames $\{I_t\}_{t=1}^{T}$; LR resolution $W\times H$; LR patch size $w\times h$; number of bins $N$
\Ensure Selected patch sets $\{P_t\}_{t=1}^{T}$

\State Define non-overlapping patches $m \leftarrow \left\lfloor \frac{W}{w} \right\rfloor\times \left\lfloor \frac{H}{h} \right\rfloor = C\times L$
\State Row-major patch index $p=r\cdot C+c,\; r\in\{1,\dots,L\},\; c\in\{1,\dots,C\}$
\State Compute $\mathbf{SF}\in\mathbb{R}^{m\times T}$ by Eq.~(\ref{eq:SF})
\State Compute $\mathbf{TF}\in\mathbb{R}^{m\times T}$ by Eq.~(\ref{eq:TF})

\For{$t = 1 \to T$}
    \State $\mathbf{e}^{(t)}_{SF} \leftarrow \textsc{Histogram}(\mathbf{SF}^{(t)}, N)$ \Comment{$|\mathbf{e}^{(t)}_{SF}|=N{+}1$}
    \State $\mathbf{C}^{(t)}_{SF} \leftarrow \textsc{Digitize}(\mathbf{SF}^{(t)}, \mathbf{e}^{(t)}_{SF}[0{:}N{-}1])$ \Comment{$\mathbf{C}^{(t)}_{SF}[p]\in\{1,\dots,N\}$}

    \If{$t=1$}
        \State $P_1 \leftarrow \{\, p \mid \mathbf{C}^{(1)}_{SF}[p]=N \,\}$ \Comment{top SF bin}
    \Else
        \State $\mathbf{e}^{(t)}_{TF} \leftarrow \textsc{Histogram}(\mathbf{TF}^{(t)}, N)$
        \State $\mathbf{C}^{(t)}_{TF} \leftarrow \textsc{Digitize}(\mathbf{TF}^{(t)}, \mathbf{e}^{(t)}_{TF}[0{:}N{-}1])$
        \State $P_t \leftarrow \{\, p \mid \mathbf{C}^{(t)}_{SF}[p]=N\ \wedge\ \mathbf{C}^{(t)}_{TF}[p]=N \,\}$ \Comment{top SF \& TF bins}
    \EndIf
\EndFor
\State \Return $\{P_t\}_{t=1}^{T}$
\end{algorithmic}
\end{algorithm}

Suppose that the resolution of a given LR video is $W\times H$, and the LR patch size is $w \times h$. The corresponding HR patch width and height are $kw$ and $kh$, where $k$ is the scaling factor. As shown in Fig.~\ref{fig:heatmap_sf_tf}, we slice each LR frame into a non-overlapping grid with $C=\lfloor W/w \rfloor$ columns and $L=\lfloor H/h \rfloor$ rows, yielding $m=C\times L$ patches per frame. When $W$ or $H$ is not divisible by $w$ or $h$, we ignore the remaining border pixels. Patch feature scores are precomputed in a vectorized manner over all $m$ patches for $T$ frames, producing per-frame score vectors $\mathbf{SF}^{(t)}\in\mathbb{R}^{m}$ and $\mathbf{TF}^{(t)}\in\mathbb{R}^{m}$ according to Eq.~(\ref{eq:SF}) and Eq.~(\ref{eq:TF}), respectively. Stacking all frames yields score matrices $\mathbf{SF}\in\mathbb{R}^{m\times T}$ and $\mathbf{TF}\in\mathbb{R}^{m\times T}$.

Given a frame $I_t$, we partition the $m$ patches into $N$ bins by applying equal-width histogram binning over the value range of $\mathbf{SF}^{(t)}$. Let $\mathbf{e}^{(t)}_{SF}\in\mathbb{R}^{N+1}$ denote the bin edges returned by the histogram operator. Each patch $p$ is assigned a bin label $\mathbf{C}^{(t)}_{SF}[p]\in\{1,\dots,N\}$ by digitizing $\mathbf{SF}^{(t)}$ using the left edges $\mathbf{e}^{(t)}_{SF}[0{:}N{-}1]$. A larger bin label corresponds to a higher SF value range. Since the binning is value-range based, different bins may contain different numbers of patches depending on the score distribution. Mathematically, the highest bin (label $N$) acts as a strict thresholding filter, capturing only the patches whose scores fall into the top interval $[\max(\mathbf{SF}^{(t)}) - \Delta_{SF}, \max(\mathbf{SF}^{(t)})]$, where the bin width is $\Delta_{SF} = (\max(\mathbf{SF}^{(t)}) - \min(\mathbf{SF}^{(t)}))/N$.

For the first frame $I_1$, the training set $P_1$ is defined as the patches in the highest SF bin, i.e., $P_1=\{p\mid \mathbf{C}^{(1)}_{SF}[p]=N\}$, which are expected to contain the most spatially complex textures. For subsequent frames $I_t$ ($2\le t\le T$), we apply the same histogram binning procedure to $\mathbf{TF}^{(t)}$ and select patches that simultaneously fall into the highest SF and TF bins:
$P_t=\{p\mid \mathbf{C}^{(t)}_{SF}[p]=N \wedge \mathbf{C}^{(t)}_{TF}[p]=N\}$. Note that in some cases, $P_t$ might be an empty set, which means that no patch in frame $I_t$ is simultaneously among the top bins of both SF and TF, thereby reducing the total number of patches. 

Finally, all the selected patches from each frame are combined as the final training set $P$. This approach saves time and computational resources and maintains model performance when no new information is available for fine-tuning. It is worth noting that although our sampling strategy does not explicitly enforce large-scale semantic coherence, this design is well-suited for content-aware SR overfitting. Unlike high-level vision tasks that rely on global semantic context, SR mainly focuses on reconstructing local textures. Since our models are fine-tuned from pre-trained networks that already capture global structural priors, focusing on informative patches allows EPS to enhance texture and motion reconstruction while preserving global coherence.

\begin{figure*}[tb]
  \centering
  \includegraphics[width=0.82\linewidth]{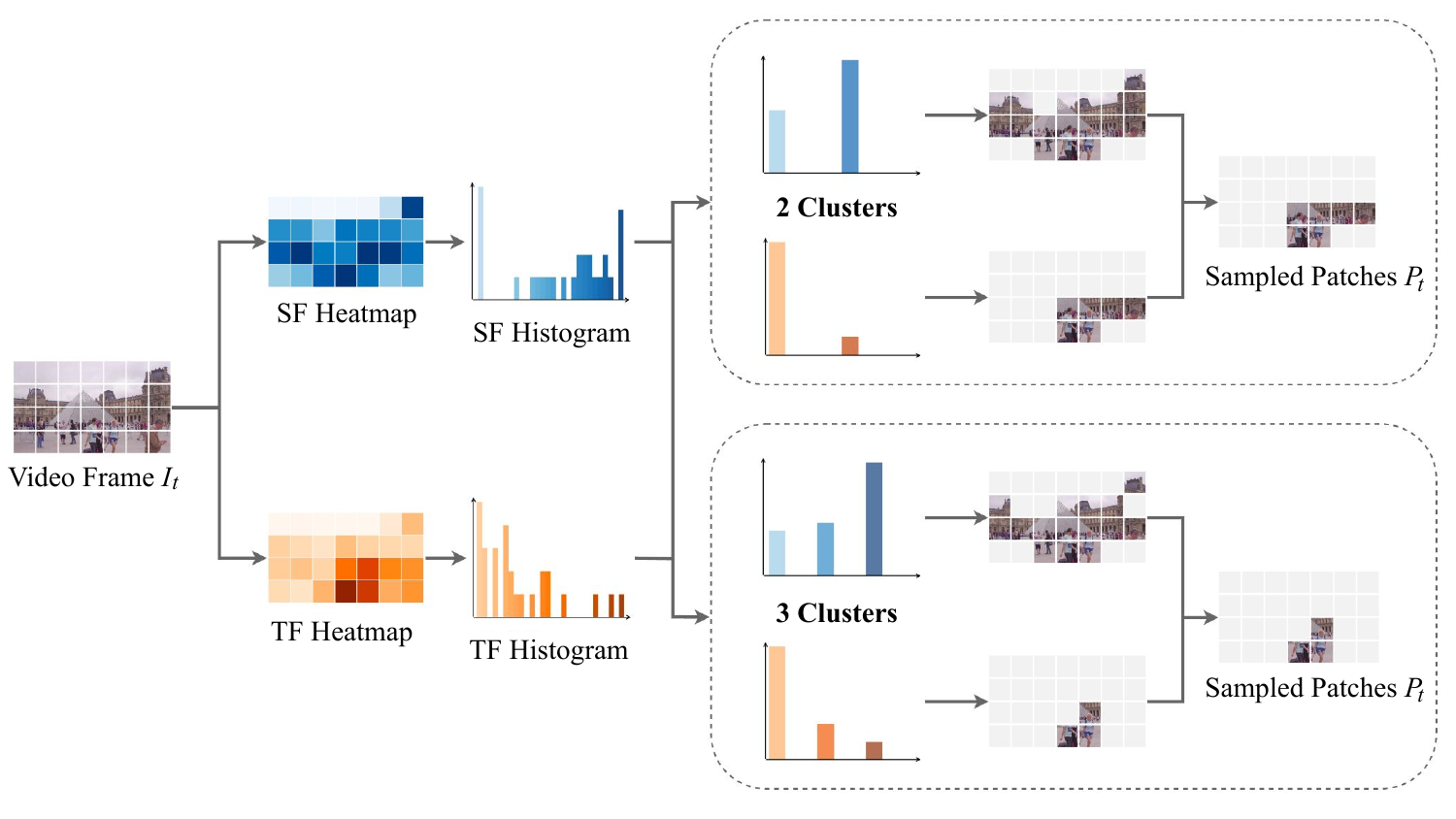}
  \caption{Example of the proposed patch sampling algorithm for $N=\{2,3\}$.
  }
  \label{fig:example_of_algorithm}
\end{figure*}

Fig.~\ref{fig:example_of_algorithm} shows an example of the proposed patch sampling algorithm for the second frame of video \#0048 from the Inter4K dataset~\cite{Inter4K_dataset}. We adjust the number of sampled patches by grouping the $SF$ and $TF$ scores into $N$ clusters, which are based on the number of bins in the histogram. Based on our empirical evaluations (detailed in Section~\ref{sec:ablation-study}), we set the default parameter to $N=2$ for our main experiments, as it provides an optimal trade-off between SR quality preservation and computational load reduction. 
Although the nominal number of clusters ($N$) is fixed, this histogram-based clustering serves as a highly dynamic and content-adaptive thresholding mechanism. Since \textit{SF} and \textit{TF} values are unnormalized and their ranges vary significantly depending on the video's texture and motion intensity, a rigid global threshold would fail to generalize. By dynamically determining the bin boundaries based on the minimum and maximum feature values of each individual frame, the actual selection threshold and the resulting number of sampled patches naturally adjust to the specific complexity of the content. Moreover, while histogram binning performs hard clustering, perceptually important transition and edge patches inherently occupy the high-value tail of the \textit{SF} and \textit{TF} distributions. This ensures they are reliably captured by the highest bin without introducing the computational overhead associated with more sophisticated algorithms.

\section{Experiments}
\label{sec:experiments}
In this section, we describe our experiments to evaluate the performance of our proposed method. Section~\ref{sec:implementation-details} comprises information about the video dataset used for our experiments and implementation details. In Section~\ref{sec:compare-with-other-methods}, we compare our EPS approach with various neural video delivery methods, including both non-sampling methods and recent patch sampling methods. Finally, in Section~\ref{sec:ablation-study}, we present our ablation study on key parameters demonstrating the impact on (\textit{i}) patch features, (\textit{ii}) clustering, (\textit{iii}) cluster number $N$, (\textit{iv}) training epoch, (\textit{v}) patch size, (\textit{vi}) quantization parameter, (\textit{vii}) and video codec. 

\subsection{Experimental Settings}
\label{sec:implementation-details}
\begin{figure}[tb]
  \centering
  \includegraphics[width=0.8\linewidth]{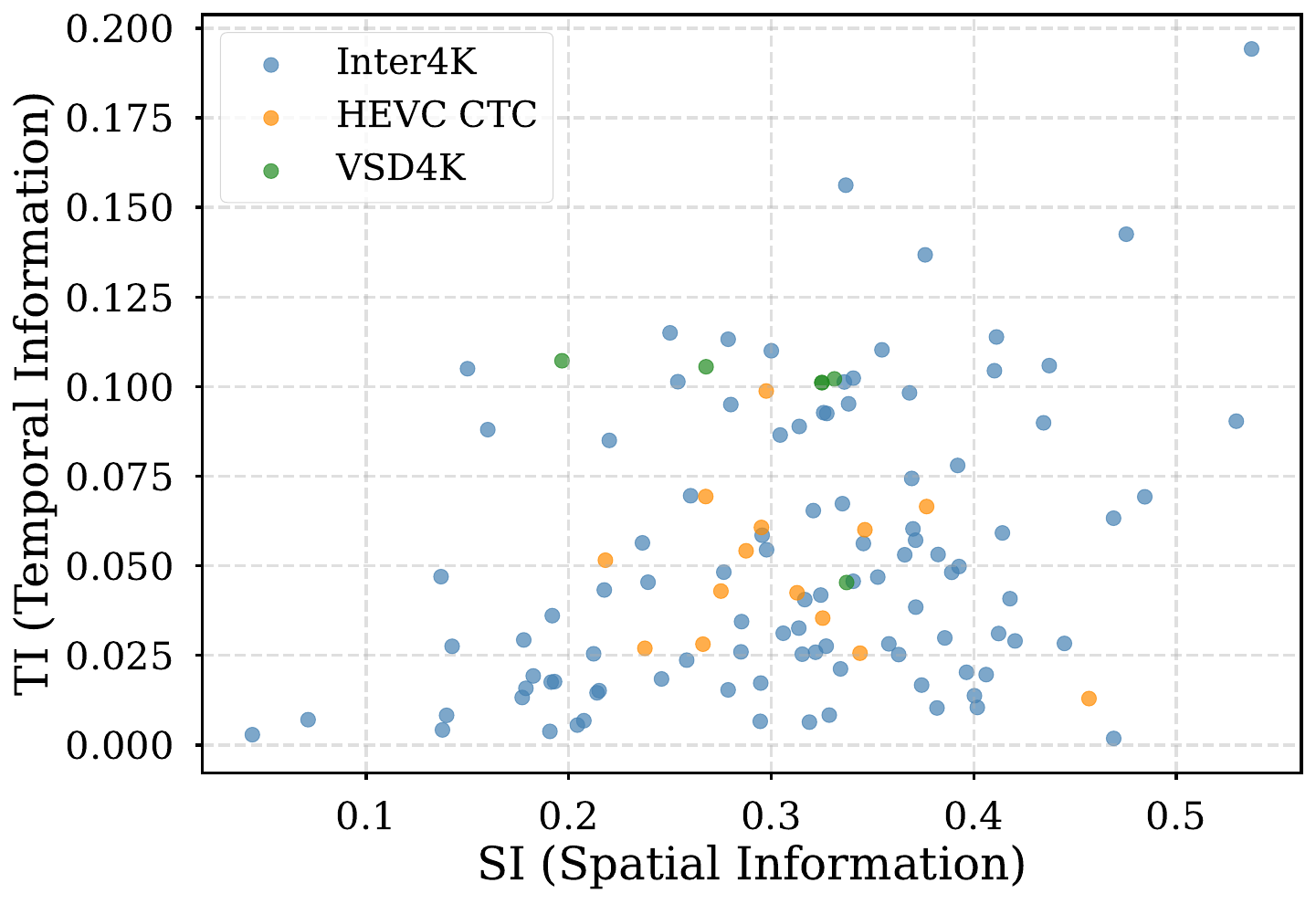}
  \caption{Scatter plot of Spatial Information (SI) versus Temporal Information (TI) for the test video benchmark.
  }
  \label{fig:si_ti_dataset}
\end{figure}

We aim to assess our method across various content complexities, thus selecting a dataset comprising diverse scenes, motions, and objects. For this study, we adopted the first 100 video sequences from the Inter4K dataset~\cite{Inter4K_dataset}, 14 video sequences from the HEVC CTC~\cite{bossen_jvet_2020} dataset, and 6 video sequences from the VSD4K~\cite{CaFM_liu_2021} dataset, to construct a comprehensive benchmark for our experiments. To quantitatively demonstrate content diversity, Fig.~\ref{fig:si_ti_dataset} plots the Spatial Information (SI) and Temporal Information (TI) of all three datasets. The broad distribution across the spatial-temporal plane confirms that our evaluation benchmarks encompass highly diverse video characteristics.
We apply bicubic downsampling to downscale the original version to a resolution of $1920\times1080$ to be used as \textit{HR video}. For \textit{LR video}, we utilize two scaling factors, i.e., $\times$2 and $\times$4, and all LR videos are compressed with four quantization parameters (QPs) values, i.e., \{22, 27, 32, 37\} using the x265 encoder\footnote{\url{https://x265.readthedocs.io/en/master/index.html}, last access: March 7, 2025.}. The set of videos with QP 27 is utilized as the default dataset. During the SR model training phase, the corresponding patches extracted from the \textit{compressed} LR videos are used as network inputs, while the HR patches serve as the ground truth. It is worth noting that for each video, we only need to compute SF and TF once using the \textit{uncompressed} LR video. To comprehensively evaluate the super-resolution (SR) performance from multiple perspectives, we adopt three established metrics: Peak Signal-to-Noise Ratio (PSNR), Learned Perceptual Image Patch Similarity (LPIPS), and Video Multi-Assessment Method Fusion (VMAF)~\cite{VMAF_2017}. All PSNR, LPIPS, and VMAF values are the average values of 120 videos. To verify the effectiveness and generalization of our method, we conduct the experiments using five SR architectures, including FSRCNN~\cite{FSRCNN_2016}, ESPCN~\cite{ESPCN_2016}, CARN~\cite{CARN_2018}, WDSR~\cite{WDSR_2018}, and SwinIR~\cite{liang2021swinir}. The LR patch size is set to 64$\times$64 during training unless mentioned otherwise. We use Adam optimizer with $\beta_1 = 0.9$, $\beta_2 = 0.999$, $\epsilon = 10^{-8}$ and we use L1 loss as loss function. The minibatch size is 64, and the learning rate is $10^{-4}$. All models are trained 300 epochs from a pre-trained model using the above experimental setups. Pre-trained models are trained using the DIV2K~\cite{DIV2K} and Flicker2K~\cite{DIV2K} datasets without any compression, but they lack training data from the Inter4K~\cite{Inter4K_dataset} dataset, making them suitable for content-aware training.

We compare our method with other neural video delivery methods, including agDNN~\cite{agDNN_2017}, where a video is super-resolved by a content-agnostic DNN (i.e., a pre-trained model without overfitting), NAS~\cite{NAS_2018} that fine-tunes the pre-trained model to fit all patches of the video without patch sampling,  LiveNAS~\cite{LiveNAS_2020} that randomly samples patches to train the content-aware model, and EMT~\cite{EMT_2022} that calculates the patch PSNR heatmap to select the $r\%$ most challenging patches of each frame to train the SR model. To assess the overfitting quality, we train content-aware SR models using the same pre-trained SR model. In the EPS method, the histogram distributions of $SF$ and $TF$ differ across video contents, leading to variations in the number of patches used for training. Here, we set the number of clusters to two in EPS and calculate the average number of training patches used across all videos. For different patch sampling methods, all training parameters and the number of training patches are kept constant, with the only difference being the selected patches. Since EPS dynamically determines the number of selected patches for each video sequence, we align all baseline methods with the average selection ratio produced by EPS, selecting the same number of top-ranked patches for each sequence to ensure a fair comparison.

For training with only small subsets of data, there are some similar approaches, such as dataset distillation~\cite{wang2018dataset_distillation} and active learning~\cite{zhang2022galaxy_activelearning}. However, these methods are not applicable to the content-aware SR model training task in this paper. First, we aim to select a small portion of data for model overfitting on video content, so it is important to preserve the original data details. Fine-tuning the model based on a synthesized dataset may not adequately address the overfitting problem and could result in suboptimal performance. Therefore, we do not compare with dataset distillation. Secondly, utilizing a deep neural network for patch selection is costly for our task. Such network training would result in additional time requirements and increased computational overhead. Identifying and labeling the impact of each patch on the model's performance is also expensive. Given a $p$-frame video with $q$ patches per frame, $2^{(p\times q)}$ training sessions are required to obtain labels. To mitigate these expenses, we only consider training-free approaches. 

To ensure a strictly fair comparison, all content-aware methods (including EPS, EMT, LiveNAS, and NAS) share the exact same training recipe and hyperparameters. The comprehensive experimental protocol is summarized in Table~\ref{tab:training_protocol}.

\setlength{\tabcolsep}{0.55mm}{\begin{table*}[t]
\small
  \centering
  \caption{Comparison of SR results between content-agnostic DNN method (agDNN~\cite{agDNN_2017}), content-aware method using all patches for training (NAS~\cite{NAS_2018}), and content-aware approach using different patch sampling approach LiveNAS~\cite{LiveNAS_2020}, EMT~\cite{EMT_2022} and EPS (ours). Results are averaged across all frames over all sequences. The best and the second best results are in \best{bold black} and \second{blue}, respectively.}
  \scalebox{0.9}{
\begin{tabular}{c|l|r|ccc|ccc|ccc|ccc|ccc}
    \toprule
        \multirow{2}*{ Scale} & \multirow{2}*{ Method} & \multirow{2}*{ Patches} & \multicolumn{3}{c|}{\bf FSRCNN} & \multicolumn{3}{c|}{\bf ESPCN} & \multicolumn{3}{c|}{\bf CARN} & \multicolumn{3}{c|}{\bf WDSR} & \multicolumn{3}{c}{\bf SwinIR} \\
      & &  & PSNR   & VMAF & LPIPS  & PSNR   & VMAF  & LPIPS & PSNR  & VMAF  & LPIPS  & PSNR    & VMAF& LPIPS & PSNR    & VMAF & LPIPS \\
  \midrule
      \multirow{5}*{$\times$ 2}  & agDNN~\cite{agDNN_2017} &  0 \% & 29.06 & 81.21  & 0.1355 & 31.81 & 80.37 & 0.1288 & 35.75 & 83.14 & 0.1229 & 35.37 & 86.27 & 0.1240 & 35.93 & 87.86 & 0.1164\\
                           & NAS~\cite{NAS_2018} & 100.00 \% & 34.53& 83.57 & 0.1193 & 33.02& 83.98 & 0.1224& \second{37.02} & \second{85.83}  & \second{0.1185} & \second{36.40} & \second{90.92} & \second{0.1128} & \second{38.47} & \second{90.34} & \second{0.0982}\\
                           &  LiveNAS~\cite{LiveNAS_2020} & 16.66 \% & 29.10 &81.42  &  0.1338    & 32.77& 81.73 & 0.1249  & 36.08 & 84.01  & 0.1219& 35.87 & 89.44 &  0.1141 &36.20 & 88.59 & 0.1107\\
                            &  EMT~\cite{EMT_2022} & 16.66 \% & \second{34.72} & \second{84.28} & \second{0.1174} & \second{33.62} & \second{84.07} & \second{0.1156} & 36.29& 84.96 & 0.1197 & 36.01& 90.46 & 0.1133 & 37.79 & 89.01 & 0.1014\\
                            &  \cellcolor{lightgray!30}\textbf{EPS (ours)} & \cellcolor{lightgray!30}16.66 \% & \cellcolor{lightgray!30}\best{34.94} & \cellcolor{lightgray!30}\best{84.36} &\cellcolor{lightgray!30}\best{0.1169} & \cellcolor{lightgray!30}\best{34.15} & \cellcolor{lightgray!30}\best{84.34} &\cellcolor{lightgray!30}\best{0.1131} & \cellcolor{lightgray!30}\best{37.19} & \cellcolor{lightgray!30}\best{86.13} &\cellcolor{lightgray!30}\best{0.1157} & \cellcolor{lightgray!30}\best{37.17} & \cellcolor{lightgray!30}\best{91.15} &\cellcolor{lightgray!30}\best{0.1117} & \cellcolor{lightgray!30}\best{38.69} &\cellcolor{lightgray!30}\best{91.24} & \cellcolor{lightgray!30}\best{0.0978}\\
    \midrule
      \multirow{5}*{$\times$ 4}  & agDNN~\cite{agDNN_2017} &  0 \% &  28.17 & 55.36 & 0.3602 & 29.01 & 38.29 & 0.4591 & 29.84 & 56.23 & 0.3879 & 29.85& 56.50 & 0.3889 & 30.86 & 56.59 & 0.3892\\
                           & NAS~\cite{NAS_2018} & 100.00 \% & 31.06 &  57.41 & 0.3568 & \second{30.58} & \second{47.51} & \second{0.4036} & \second{32.17} & \second{64.76} & \second{0.3030} & \best{32.06} & \best{65.59} & \best{0.3025} & \best{32.25} & \best{71.24} & \best{0.2635}\\
                           &  LiveNAS~\cite{LiveNAS_2020} & 25.00 \% & 30.91 & 55.97 & 0.3583 & 30.29 & 43.62 & 0.4284 & 31.26 & 59.37 & 0.3481 & 31.17 & 60.36 & 0.3517 & 31.25 & 62.36 & 0.3587 \\
                           &  EMT~\cite{EMT_2022} & 25.00 \% & \second{31.07} & \second{57.54} & \second{0.3565} & 30.42 & 46.81 & 0.4137 & 32.03 & 62.99 & 0.3144 & 31.64 & 62.23 & 0.3180 & 31.98 & 68.55 & 0.2803\\
                            &  \cellcolor{lightgray!30}\textbf{EPS (ours)}  & \cellcolor{lightgray!30}25.00 \% &\cellcolor{lightgray!30}\best{31.23} & \cellcolor{lightgray!30}\best{57.67} & \cellcolor{lightgray!30}\best{0.3563} & \cellcolor{lightgray!30}\best{30.63} & \cellcolor{lightgray!30}\best{47.53} & \cellcolor{lightgray!30}\best{0.4025} & \cellcolor{lightgray!30}\best{32.24} & \cellcolor{lightgray!30}\best{64.80} & \cellcolor{lightgray!30}\best{0.2981} & \cellcolor{lightgray!30}\second{31.93} & \cellcolor{lightgray!30}\second{64.94} & \cellcolor{lightgray!30}\second{0.3036} & \cellcolor{lightgray!30}\second{32.18} & \cellcolor{lightgray!30}\second{70.31} & \cellcolor{lightgray!30}\second{0.2696}\\
  \bottomrule
  \end{tabular}
}
  \label{tab:compare_with_model_all_patches}
\end{table*}
}

\begin{table}[tb]
\centering
\caption{Experimental protocol for all compared methods.}
\label{tab:training_protocol}
\footnotesize
\begin{tabular}{@{} l p{0.55\columnwidth} @{}}
\toprule
\textbf{Setting} & \textbf{Configuration} \\
\midrule
\multicolumn{2}{@{}l}{\textit{Optimization Parameters}} \\
\midrule
Initialization & Identical pre-trained model weights \\
Optimizer & Adam ($\beta_1=0.9$, $\beta_2=0.999$, $\epsilon=10^{-8}$) \\
Learning Rate & $10^{-4}$ \\
Loss Function & L1 Loss \\
Batch Size & 64 \\
Training Epochs & 300 \\
\midrule
\multicolumn{2}{@{}l}{\textit{Data \& Evaluation Parameters}} \\
\midrule
Random Seed & 42 \\
Data Augmentations & None (to isolate spatial-temporal feature impacts) \\
Quantization Parameter & 27 (Default) \\
LR Patch Size & $64 \times 64$ \\
Patch Count ($\times 2$) & Average 600 of 3600 patches per sequence (16.66\%) \\
Patch Count ($\times 4$) & Average 210 of 840 patches per sequence (25.00\%) \\
\bottomrule
\end{tabular}
\end{table}

\begin{figure*}[t!]
   \centering
  \includegraphics[width=0.96\linewidth]{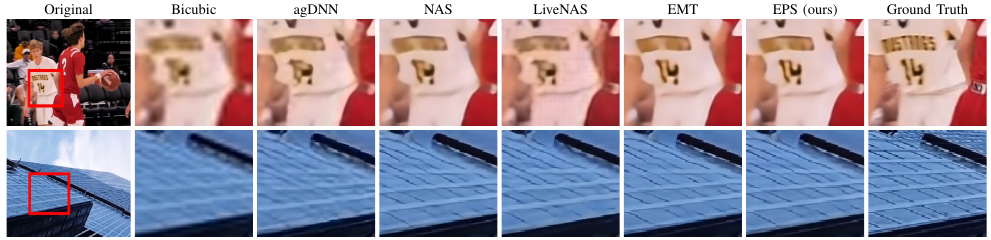}
  \caption{Super-resolution quality comparison using our method (seventh column) with baseline methods. }
  \label{fig:visual_comparison}
\end{figure*}

\begin{table*}[!h]
\linespread{0.9}
   \small
  \centering
    \caption{Impact of cluster number. The Patches column denotes the average portion of patches utilized for training.
  }
    \scalebox{0.8}{\begin{tabular}{@{}c|c|r|ccc|ccc|ccc|ccc|ccc@{}}
    \toprule
     \multirow{2}*{\bf Scale Factor} & \multirow{2}*{\bf \#Clusters} & \multirow{2}*{\bf Patches} & \multicolumn{3}{c|}{\bf FSRCNN} & \multicolumn{3}{c|}{\bf ESPCN} & \multicolumn{3}{c|}{\bf CARN} & \multicolumn{3}{c|}{\bf WDSR} & \multicolumn{3}{c}{\bf SwinIR}\\
      & &  & PSNR   & VMAF  & LPIPS & PSNR   & VMAF & LPIPS & PSNR   & VMAF & LPIPS & PSNR   & VMAF  & LPIPS& PSNR   & VMAF & LPIPS \\
 \midrule
      \multirow{5}*{$\times$ 2}  & 1 & 100.00 \% & \second{34.64}& \second{83.57} & \second{0.1186} & 33.02& 83.98 & 0.1224& \second{37.02} & \second{85.83}  & \second{0.1185} & \second{36.40} & \second{90.92} & \second{0.1128} & \second{38.47} & \second{90.34} & \second{0.0982}\\
                           &  2 & 16.66 \% & \cellcolor{lightgray!30}\best{34.94} & \cellcolor{lightgray!30}\best{84.36} &\cellcolor{lightgray!30}\best{0.1169} & \cellcolor{lightgray!30}\best{34.15} & \cellcolor{lightgray!30}\best{84.34} &\cellcolor{lightgray!30}\best{0.1131} & \cellcolor{lightgray!30}\best{37.19} & \cellcolor{lightgray!30}\best{86.13} &\cellcolor{lightgray!30}\best{0.1157} & \cellcolor{lightgray!30}\best{37.17} & \cellcolor{lightgray!30}\best{91.15} &\cellcolor{lightgray!30}\best{0.1117} & \cellcolor{lightgray!30}\best{38.69} &\cellcolor{lightgray!30}\best{91.24} & \cellcolor{lightgray!30}\best{0.0978}\\
                           &  3 & 8.31 \% & 34.53 & 83.45 & 0.1193& \second{34.02} & \second{84.12} & \second{0.1175} & 37.01  & 85.94 & 0.1163  & 36.41  & 90.72 & 0.1125 & 38.32 & 90.08 & 0.0991\\
                           &  4 & 5.83 \% & 34.43 & 83.29 & 0.1203 & 33.97 & 84.08  & 0.1194 & 37.96 & 85.31 & 0.1192&36.22 & 89.45 & 0.1134& 38.27 & 89.13 & 0.0993 \\
                           &  5 & 3.24 \% & 34.16 & 83.12 & 0.1215& 33.17 & 83.84 & 0.1218 & 36.88 & 85.67 & 0.1192& 36.21 & 89.76 &  0.1149 & 38.14 & 88.74 & 0.0998\\
      \midrule
      \multirow{5}*{$\times$ 4}  & 1 & 100.00 \%&31.06 &  57.41 & 0.3568 & \second{30.58} & \second{47.51} & \second{0.4036} & \second{32.17} & \second{64.76} & \second{0.3030} & \best{32.06} & \best{65.59} & \best{0.3025} & \best{32.25} & \best{71.24} & \best{0.2635}\\
                           &  2 &  25.00 \% &\cellcolor{lightgray!30}\best{31.23} & \cellcolor{lightgray!30}\best{57.67} & \cellcolor{lightgray!30}\best{0.3563} & \cellcolor{lightgray!30}\best{30.63} & \cellcolor{lightgray!30}\best{47.53} & \cellcolor{lightgray!30}\best{0.4025} & \cellcolor{lightgray!30}\best{32.24} & \cellcolor{lightgray!30}\best{64.80} & \cellcolor{lightgray!30}\best{0.2981} & \cellcolor{lightgray!30}\second{31.93} & \cellcolor{lightgray!30}\second{64.94} & \cellcolor{lightgray!30}\second{0.3036} & \cellcolor{lightgray!30}\second{32.18} & \cellcolor{lightgray!30}\second{70.31} & \cellcolor{lightgray!30}\second{0.2696}\\
                           &  3 & 17.85 \%& \second{31.15} & \second{57.47} & \second{0.3567}  & 29.87 & 47.37 & 0.4101 &  31.78 & 64.51 & 0.3064  & 31.92 & 64.24 & 0.3042& 32.07 & 68.73 & 0.2849 \\
                           &  4 & 10.71 \% & 30.92 & 56.64 & 0.3572 & 29.62 & 47.06 & 0.4128 &31.20  & 63.67 & 0.3129 &31.34  & 63.97 & 0.3067& 31.66 & 67.48 & 0.2931\\
                           &  5 & 7.14 \% & 30.76 & 56.28 & 0.3585 & 29.34 & 46.81  & 0.4139&30.97 & 63.55 &0.3226& 31.27 & 63.81 & 0.3108& 31.37 & 65.29 & 0.3206\\
  \bottomrule
  \end{tabular}
  }
  \label{tab:ablation_study_number_cluster}
\end{table*}

\subsection{Comparison with Related Works}
\label{sec:compare-with-other-methods}

Table~\ref{tab:compare_with_model_all_patches} shows a comprehensive quantitative comparison with various SR models under different scaling factors. As can be seen, EPS consistently achieves similar or even higher quality performance compared to the NAS trained with all available patches. EPS also demonstrates improved performance over other patch sampling methods (i.e., LiveNAS and EMT) using the same number of patches. The qualitative comparison is shown in Fig.~\ref{fig:visual_comparison}.

For the $\times$2 scaling factor, the LR video has a resolution of $960\times540$. We partition each 30-frame video sequence into a total of 3,600 patches, which equals 120 patches per frame. Through our patch sampling algorithm in Section~\ref{sec:patch-sampling}, we select an average of 600 out of 3600 patches per sequence (i.e., 16.66\%) for content-aware training, resulting in similar or even slightly better training gains compared to NAS. This might be because training NAS over non-informative patches reduces the model's overall effectiveness. In the case of the $\times$4 scaling factor, the LR video has a resolution of $480\times270$, resulting in 840 patches for a 30-frame video sequence. The overall patch count is reduced from 840 to an average of 210 (i.e., 25\%), using the proposed patch selection algorithm. The quality achieved by the content-aware DNN, measured by PSNR, VMAF, and LPIPS, is significantly superior to the content-agnostic DNN (agDNN) for NAS and EPS. In some cases, there is even an improvement in EPS compared to NAS, which uses all patches for training. Conversely, models with larger parameter sizes, like WDSR and SwinIR, exhibit a moderate decrease in quality at the $\times$4 scale. However, EPS can highly reduce the training data of the model with only a negligible drop in quality compared to NAS.

\begin{figure}[!h]
  \centering
  \includegraphics[width=0.99\linewidth]{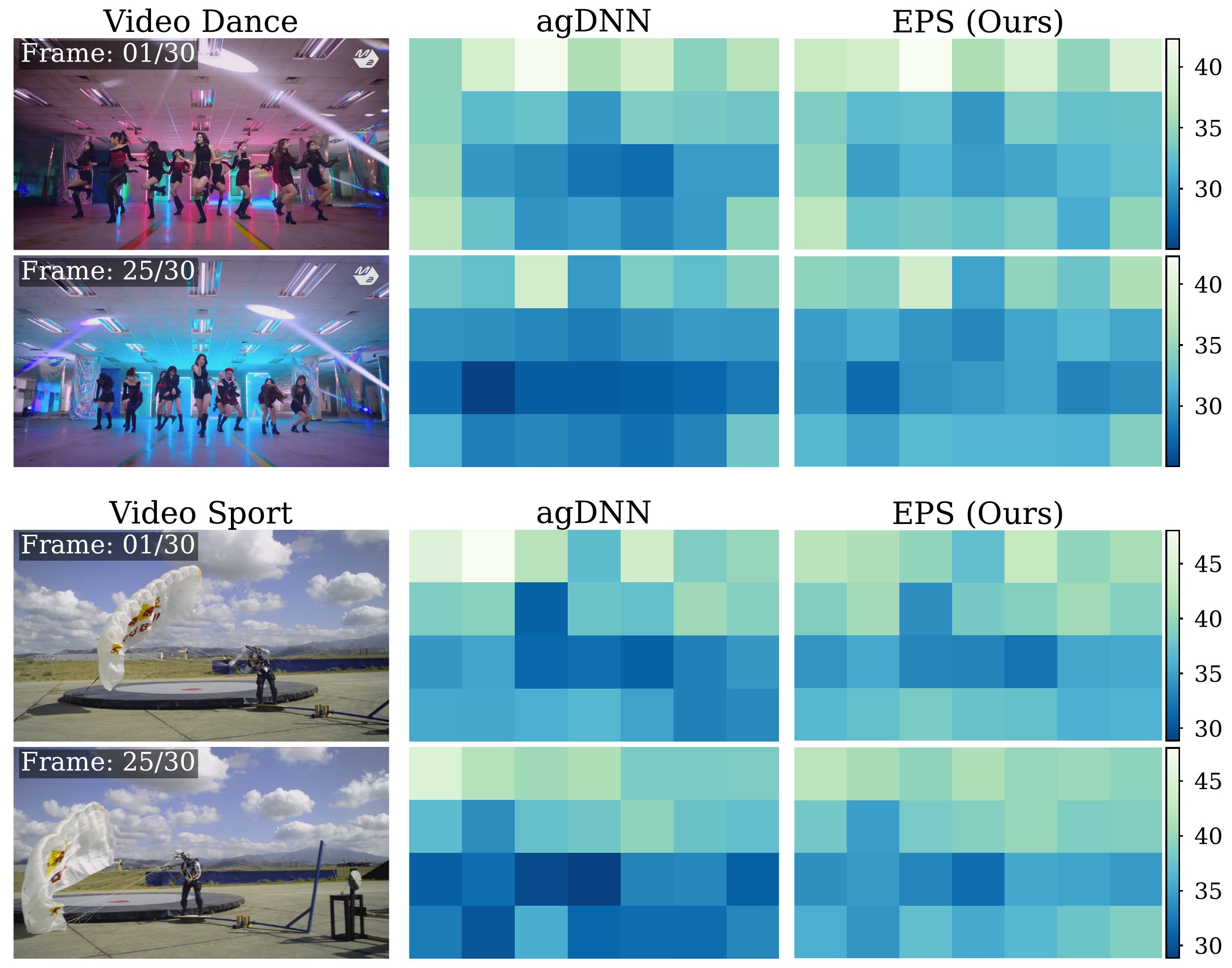}
  \caption{Patch PSNR heatmaps of $\times$4 video SR using the CARN model. Compared with the pre-trained agDNN baseline, our EPS method achieves noticeable improvements in complex regions with fine textures and motion.
  }
  \label{fig:patch_heatmap_dance_sport}
\end{figure}

To illustrate where the performance gains occur spatially within a sequence, we plot patch PSNR heatmaps for $\times4$ SR videos using the CARN model, comparing the pre-trained agDNN baseline with our EPS fine-tuning method. As shown in Fig.~\ref{fig:patch_heatmap_dance_sport}, we present the \textit{Dance} and \textit{Sport} sequences from the VSD4K dataset. The heatmaps show that our method achieves higher PSNR in complex regions with fine textures and strong motion. In contrast, less complex regions such as flat backgrounds already exhibit high PSNR before fine-tuning, leading to only marginal gains. This observation verifies that the EPS framework effectively selects spatial-temporal complex patches and focuses improvements on key content regions that contain important visual structures.

\begin{figure}[tb]
  \centering
  \includegraphics[width=0.88\linewidth]{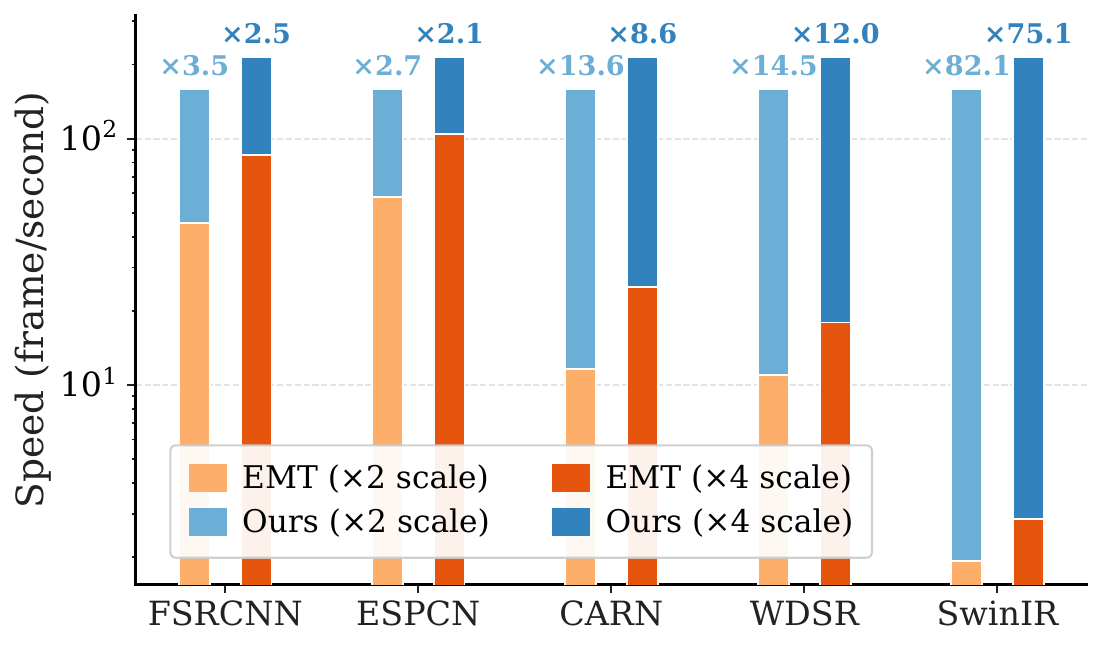}
  \caption{Comparison of the average patch sampling speed (frame/second) between EMT and our EPS.
  }
  \label{fig:comparison_speed}
\end{figure}

LiveNAS has a fast sampling speed due to the random selection of patches, but the training gain is significantly lower than that of EPS trained with all patches. While our approach demonstrates better performance compared to EMT, it is equally crucial to assess the time required during patch selections. We measured the average execution speed for sampling the same number of LR-HR patches from 120 videos, as shown in Fig.~\ref{fig:comparison_speed}. The results are averaged over 10 runs and evaluated on an NVIDIA RTX 6000 Ada GPU. Our approach achieves a speedup of up to 82.1$\times$ in patch sampling time compared to EMT. The disparity in run time is linked to the parallel-friendly and model-agnostic design of our approach, where the computations of $SF$ and $TF$ can be performed concurrently. In general, our EPS method for the same number of selected patches achieves higher quality at a significantly reduced run time.

We also analyze a training time breakdown in Fig.~\ref{fig:video_genre_texture_motion}, where the patch sampling process is included in the total training time. The results show that EPS consistently reduces the overall training time across all models. Even for smaller models such as FSRCNN and ESPCN, the training time is reduced by about 52.0\%. The acceleration becomes more evident as the model size increases. For larger models, including CARN, WDSR, and SwinIR, the training time reduction ranges from 61.5\% to 88.7\%.  Overall, these results demonstrate that EPS reliably reduces the training cost across models of different scales. Meanwhile, the patch sampling process itself accounts for only a negligible fraction of the total training time, showing that the overhead introduced by the feature computation is negligible.

\begin{figure}[tb]
  \centering
  \includegraphics[width=0.88\linewidth]{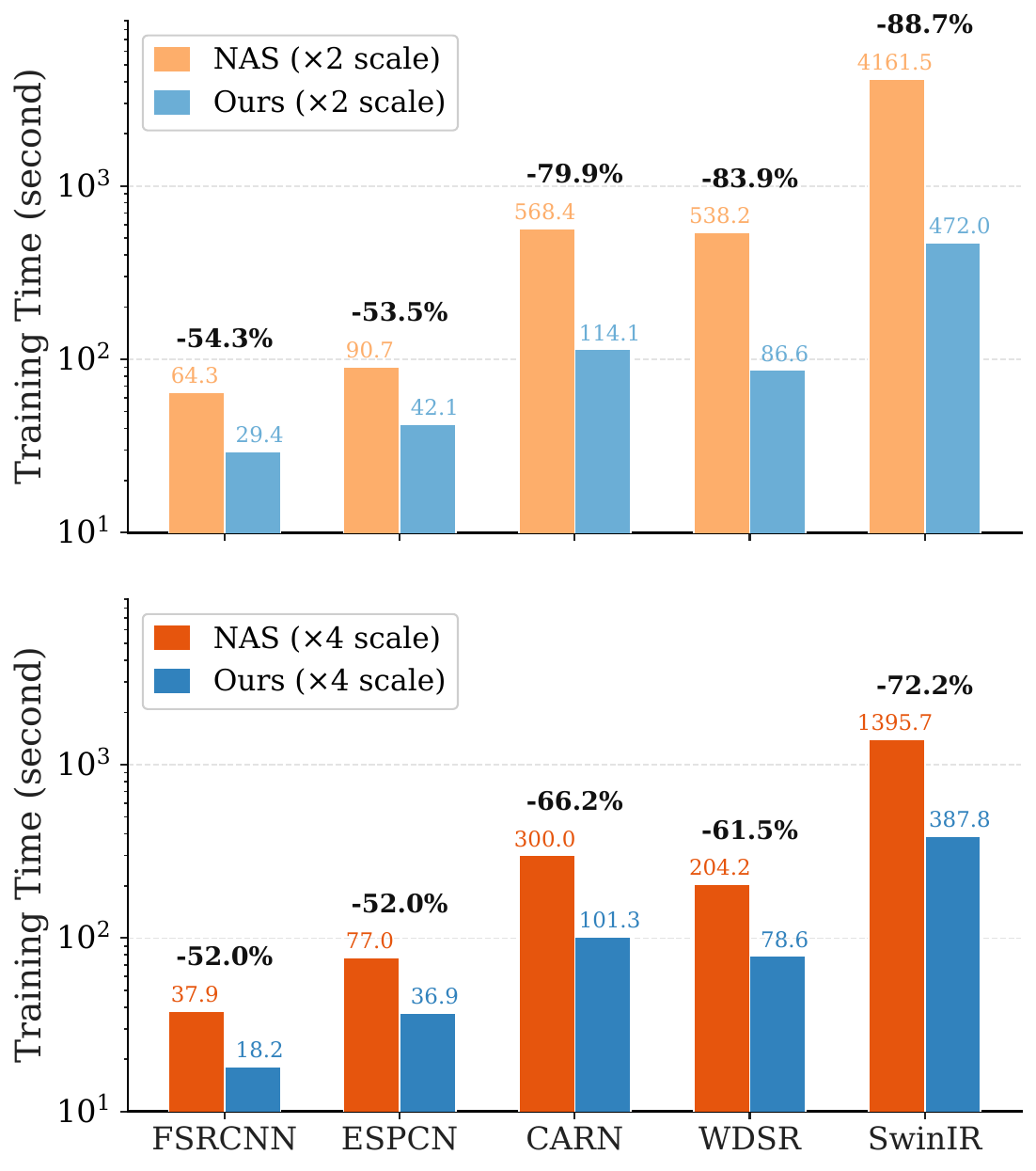}
  \caption{Training time comparison between full data training (NAS) and our EPS under the same experimental protocol. The EPS patch sampling process accounts for only 0.23s ($\times$2 scale) and 0.26s ($\times$4 scale) of the total training time.}
  \label{fig:video_genre_texture_motion}
\end{figure}

\subsection{Ablation Study}
\label{sec:ablation-study}

% \noindent\textbf{$\rhd$ Impact of Patch Features.}
% We compare the performance of different patch features using Sobel edge detection and DCT-based energy function. We replaced the proposed weighted DCT values with the mean and std of the Sobel filter of patches ($\text{Sobel}_{\text{mean}} $, $ \text{Sobel}_{\text{std}}$), and selected the highest $r\%$ complex patches using different patch features. The SR performance decreased as shown in Table~\ref{tab:compare_feature_map}.  Both methods yield spatial-temporal complexity features, but DCT-based features ({\it SF} and {\it TF}) generally perform better in video compression due to the reliance of codecs on DCT. 
% A key advantage of the proposed method is its potential integration with the encoding process, utilizing  DCT calculations in codec to accelerate patch selection. 

\noindent $\rhd$ \textbf{Impact of Patch Features.} We compare the performance of different patch evaluation features, including efficient model-free metrics (Sobel edge detection and our DCT-based energy function) and recent deep learning-based feature selectors (IPS~\cite{bergner2022ips}, AgentViT~\cite{cauteruccio2025rl_vit}, and CAMixerSR~\cite{wang2024camixersr}). To ensure a fair comparison, we applied a fixed threshold of $r\%$ to guarantee that every method selects the exact same number of top-ranked complex patches (e.g., 25.00\% at the $\times$4 scale) based on their individual scoring mechanisms. 

As shown in Table~\ref{tab:compare_feature_map}, traditional model-free features, such as the mean and standard deviation of Sobel filters, are extremely fast to compute but result in a noticeable decrease in SR performance. Conversely, recent learning-based features achieve competitive quality improvements but introduce significant computational overhead during the selection phase due to the heavy burden of neural network inference. Furthermore, these learning-based methods require extensive pre-training and lack flexibility; adapting them to different input image resolutions or dynamic selection thresholds often requires modifying the network architecture and retraining.

In contrast, our proposed EPS achieves superior SR quality with negligible selection cost, as it relies on low-complexity, analytical DCT operations. For instance, EPS is nearly 11$\times$ faster than CAMixerSR in selection speed while maintaining or exceeding its SR quality. This validates that EPS provides a superior trade-off between training efficiency and model performance. While both Sobel and DCT yield spatial-temporal complexity features, DCT-based features (\textit{SF} and \textit{TF}) generally perform better in video compression tasks due to the inherent reliance of video codecs on DCT. A key advantage of the proposed method is its potential integration with the encoding process, utilizing DCT calculations already present in the codec to further accelerate patch selection without additional overhead.

\begin{table*}[h]
  \centering
  \caption{Quantitative comparison of SR quality and selection efficiency at $\times$4 scale. The table contrasts our training-free DCT method with Sobel filter and computation-heavy learning-based methods, utilizing the same number of training patches.}
  \scalebox{0.95}{
  \begin{tabular}{l|c|ccc|ccc|ccc|ccc|ccc}
    \toprule
    \multirow{2}*{\bf  Selector} & \bf Speed  & \multicolumn{3}{c|}{\bf FSRCNN} & \multicolumn{3}{c|}{\bf ESPCN} & \multicolumn{3}{c|}{\bf CARN} & \multicolumn{3}{c|}{\bf WDSR} & \multicolumn{3}{c}{\bf SwinIR} \\
     &  (per frame)   & PSNR   & VMAF & LPIPS  & PSNR   & VMAF  & LPIPS & PSNR  & VMAF  & LPIPS  & PSNR    & VMAF& LPIPS & PSNR    & VMAF & LPIPS \\
    \midrule
    \multicolumn{17}{c}{\textit{Deep Learning-based Feature}} \\
    \midrule
    IPS~\cite{bergner2022ips} & 98ms  & 31.02 & 57.20 & 0.3571& 30.47 & 47.14 &0.4118 & 31.82 & 62.89& 0.3120 & 31.58 & 61.62 & 0.3197& 32.03 & 68.58 &0.2801\\
    AgentViT~\cite{cauteruccio2025rl_vit}  & 74ms& \second{31.18} & 57.59  &0.3568& \second{30.59} & \second{47.33} & \second{0.4056}& \second{32.21} & \second{63.83} & \second{0.3106} & 31.75 & 63.77 & 0.3099& \second{32.04} & \second{69.23} & \second{0.2737}\\
    CAMixerSR~\cite{wang2024camixersr}& 67ms  & 31.16 & \second{57.61} &\second{0.3565} & 30.54 & 47.27 &0.5051& 32.16 & 63.74 & 0.3114& \second{31.79} & \second{64.06} & \second{0.3083} & 31.97 & 68.36 &0.2824 \\
    \midrule
    \multicolumn{17}{c}{\textit{Efficient Model-Free Feature}} \\
    \midrule
    {\bf $\text{Sobel}_{\text{mean}}$} & 6ms  & 30.76 & 56.51 & 0.3581 & 30.32 & 44.53 & 0.4209 &31.61 & 61.54& 0.3285 & 31.26 & 61.41 & 0.3277& 31.48 & 64.76 & 0.3172\\
    {\bf $\text{Sobel}_{\text{std}}$} & 6ms  & 30.94 & 56.13 & 0.3576 &  30.40 & 44.71 & 0.4176 &31.74 & 61.61 &0.3259 & 31.58 & 61.88 & 0.3293& 31.70 & 66.98 & 0.3053\\
     \cellcolor{lightgray!30}\textbf{DCT (ours)} & \cellcolor{lightgray!30}6ms &\cellcolor{lightgray!30}\best{31.23} & \cellcolor{lightgray!30}\best{57.67} & \cellcolor{lightgray!30}\best{0.3563} & \cellcolor{lightgray!30}\best{30.63} & \cellcolor{lightgray!30}\best{47.53} & \cellcolor{lightgray!30}\best{0.4025} & \cellcolor{lightgray!30}\best{32.24} & \cellcolor{lightgray!30}\best{64.80} & \cellcolor{lightgray!30}\best{0.2981} & \cellcolor{lightgray!30}\best{31.93} & \cellcolor{lightgray!30}\best{64.94} & \cellcolor{lightgray!30}\best{0.3036} & \cellcolor{lightgray!30}\best{32.18} & \cellcolor{lightgray!30}\best{70.31} & \cellcolor{lightgray!30}\best{0.2696}\\
  \bottomrule
  \end{tabular}
  }
  
  \label{tab:compare_feature_map}
\end{table*}

\noindent\textbf{$\rhd$ Impact of Clustering.} 
\label{sec:clustering} 
To evaluate the effectiveness of our dynamic clustering approach, we compare it against a baseline that selects a fixed percentage of patches for each video. We report the average SR performance improvements over the agDNN baseline across all models. As shown in Table~\ref{tab:compare_fixed_threshold}, our clustering strategy achieves higher SR performance than sampling a fixed number of patches for EPS. Furthermore, to isolate the benefits of the clustering algorithm from our feature extraction, we applied our dynamic clustering strategy to the PSNR-based feature maps generated by EMT. While this modification noticeably improved EMT's performance over its default fixed-threshold approach, it still remained inferior to our complete EPS method. As discussed in Section~\ref{sec:patch-sampling}, this improvement occurs for clustering because it inherently accounts for content dependency by dynamically adapting the number of selected patches to the video's spatial-temporal complexity. Consequently, more patches are sampled for highly complex videos, whereas fewer are retained for simpler contents, effectively reducing the training data.

\begin{table}[t]
  \centering
  \caption{Comparison of fixed threshold and clustering by sampling the same number of patches.}
    \scalebox{0.95}{\begin{tabular}{@{}c|l|ccc|ccc@{}}
    \toprule
     \multirow{2}*{Feature} & \multirow{2}*{Strategy}  & \multicolumn{3}{c|}{\bf $\times$2} & \multicolumn{3}{c}{\bf $\times$4}\\
     &  & {\bf $\Delta$PSNR}   & {\bf $\Delta$VMAF} & {\bf $\Delta$LPIPS}        & {\bf $\Delta$PSNR}   & {\bf $\Delta$VMAF}  & {\bf $\Delta$LPIPS} \\
    \midrule
             \multirow{2}*{EMT} & Fixed &  $+$2.10 &$+$2.78 &  $-$0.0120 & $+$1.88 & $+$7.03 & $-$0.0605 \\
             & Cluster &  $+$2.24 &$+$2.93 &  $-$0.0132 & $+$1.93 & $+$7.95 & $-$0.0628 \\
    \midrule
            \multirow{2}*{EPS} & Fixed  & $+$2.16 &$+$3.02 &  $-$0.0136 & $+$1.97 & $+$8.14 & $-$0.0681 \\
              & \cellcolor{lightgray!30}Cluster& \cellcolor{lightgray!30}$+$2.84 &\cellcolor{lightgray!30}$+$3.67 &  \cellcolor{lightgray!30}$-$0.0145 &\cellcolor{lightgray!30}$+$2.11 & \cellcolor{lightgray!30}$+$8.45 & \cellcolor{lightgray!30}$-$0.0710 \\
  \bottomrule
  \end{tabular}}
  \label{tab:compare_fixed_threshold}
\end{table}

\noindent\textbf{$\rhd$ Impact of Cluster Number.}
We conduct extensive experiments to explore the content-aware SR performance under different numbers of clusters during patch sampling, as shown in Table~\ref{tab:ablation_study_number_cluster}. By dynamically adjusting the number of clusters, we achieve a reduction in the number of training patches ranging from 75.00\% to 91.69\% while maintaining quality. For an SR model with a very small number of parameters, like FSRCNN and ESPCN, even a limited patch selection can yield promising results. However, for CARN, WDSR, and SwinIR, as the number of clusters separated by histogram bins increases, indicating a minimal portion of the selected patches, the trained SR model exhibits a slight quality drop. As further detailed in the table, setting $N=2$ successfully mitigates this drop and achieves an optimal balance. It significantly reduces the patch count (to 16.66\% and 25.00\% for $\times 2$ and $\times 4$ scales, respectively) while preserving high SR performance, directly justifying our choice of $N=2$ as the default configuration for the main experiments in Section~\ref{sec:compare-with-other-methods}.

\begin{figure*}[t]
  \centering
  \includegraphics[width=0.99\linewidth]{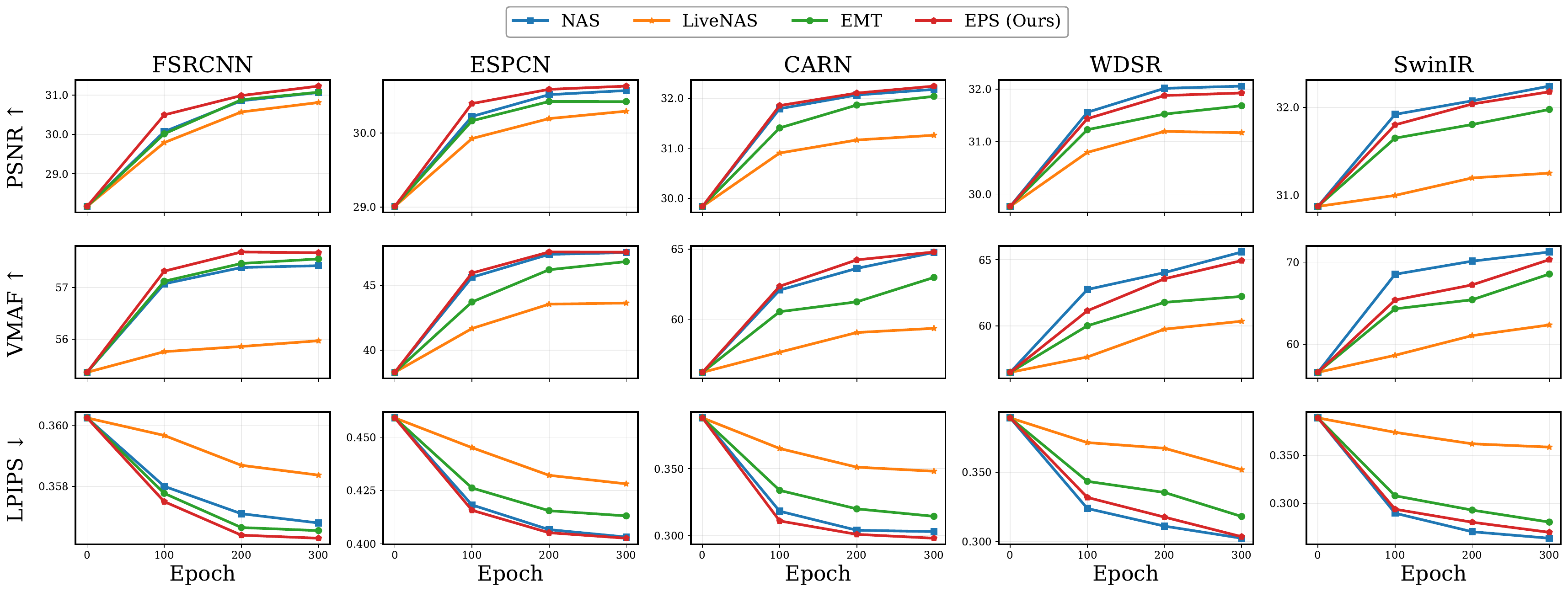}
  \caption{Comparison of video quality for all models ($\times 4$) between content-aware models when all patches are selected for training (NAS) and when different sampling approaches are utilized (LiveNAS, EMT, and EPS) for every 100 epochs.}
  \label{fig:training_epoch}
  \vspace{-1em}
\end{figure*}

\noindent\textbf{$\rhd$  Impact of Training Epoch.} We intend to explore how adjusting the training epochs affects the performance of SR models, given that content-aware training takes advantage of the overfitting property of DNNs. We train all models from pre-trained initialization to 300 epochs in scaling factor $\times4$ for every 100 epochs across all videos, using different patch sampling methods. As can be seen in Fig.~\ref{fig:training_epoch}, compared to NAS that trains with all patches, models trained with our sampled patches achieve better SR video quality with similar or much lower training epochs. It indicates that when all patches are selected for training, the model undergoes fine-tuning for both informative and non-informative patches, leading to lower performance compared to when only informative patches are utilized for training. Compared to other patch sampling methods like LiveNAS and EMT, our EPS method gets significantly higher training gains using the same number of sampled patches. This indicates that EPS is more effective at selecting informative patches, allowing the content-aware SR model to learn from higher-quality data.

\noindent $\rhd$ \textbf{Impact of Video Content.} To demonstrate the robustness of the proposed EPS framework across diverse video genres, motions, and textures, we further analyze its performance from a content-oriented perspective.

First, to validate stability across different genres, we evaluate individual sequences from the VSD4K dataset, including six representative videos that cover distinct thematic categories. Second, to assess the impact of textures and motions, we quantitatively categorize our test sequences based on their Spatial Information (SI, representing texture complexity) and Temporal Information (TI, representing motion intensity). By applying an SI threshold of $0.35$ and a TI threshold of $0.75$, we divide the dataset into four distinct spatial-temporal quadrants and evaluate the $\times$4 scale SR performance within each category.

As illustrated in Fig.~\ref{fig:video_genre_texture_motion}(a), despite drastic variations in video content, EPS consistently achieves superior SR quality improvements compared to the content-agnostic baseline (agDNN). Notably, EPS maintains this high performance while utilizing only a fraction of the training data, proving its consistency across various genres. Furthermore, the results in Fig.~\ref{fig:video_genre_texture_motion}(b) demonstrate that EPS dynamically adapts to the intrinsic distribution of the video, regardless of whether it is dominated by dense high-frequency textures or intense motion dynamics. Even for highly challenging videos with both complex textures (high SI) and intense motions (high TI), fine-tuning with our patch sampling method yields consistent and often greater quality improvements.

\begin{figure*}[tb]
  \centering
  \includegraphics[width=0.68\linewidth]{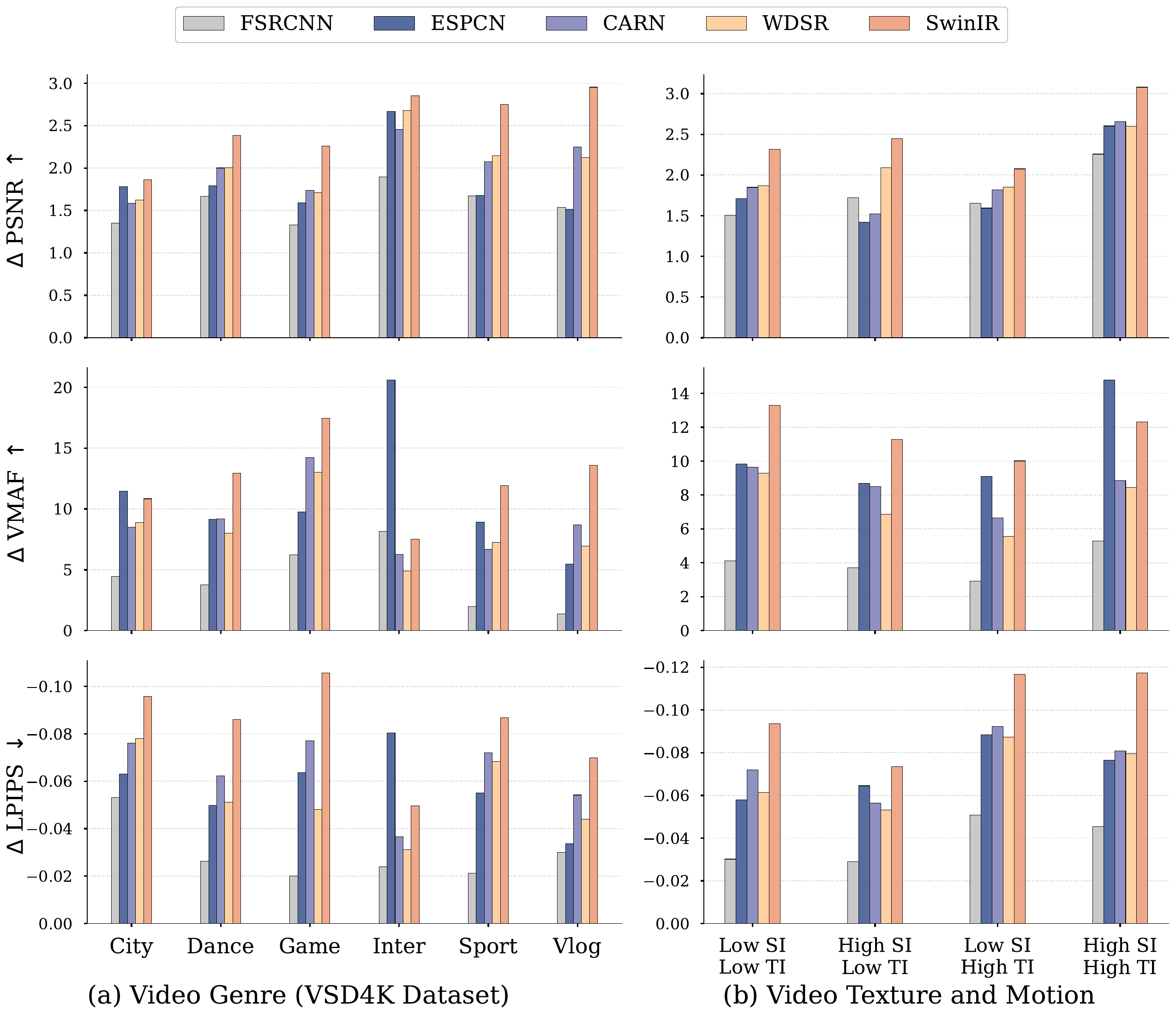}
  \caption{SR performance gain of EPS compared with the agDNN initialization across different video genres and SI–TI categories.}
  \label{fig:video_genre_texture_motion}
\end{figure*}

\noindent\textbf{$\rhd$ Impact of Patch Size.}
During the experiments, we slice each LR frame into patches of $64\times64$ by default. We also study the effect of different patch sizes. We evaluated all models using the sampling method through a two-cluster histogram and calculated their training gains based on the pre-trained initialization. The result for each patch size is reported as the average training gains of five models. As shown in Table~\ref{tab:ablation_study_pacth_size}, all patch sizes achieve similar results with negligible quality differences, showcasing that our sampling method remains highly robust and adaptable to various patch sizes.

\begin{table}[t]
  \centering
  \caption{Variants of patch size.  }
  \begin{tabular}{@{}c|ccc|ccc@{}}
    \toprule
     \multirow{2}*{\bf Patch Size} & \multicolumn{3}{c|}{\bf $\times$2} & \multicolumn{3}{c}{\bf $\times$4}\\
     & {\bf $\Delta$PSNR}   & {\bf $\Delta$VMAF} & {\bf $\Delta$LPIPS}        & {\bf $\Delta$PSNR}   & {\bf $\Delta$VMAF}  & {\bf $\Delta$LPIPS} \\
    \midrule
             64$\times$64 & $+$2.84 & $+$3.67 &  $-$0.0145 & $+$2.11 & $+$8.45 & $-$0.0712 \\
             32$\times$32&  $+$2.73 & $+$3.66  & $-$0.0136 & $+$2.06 &  $+$8.41  &  $-$0.0710 \\
             16$\times$16 & $+$2.85 & $+$3.69  & $-$0.0149 & $+$2.08 &  $+$8.50  &  $-$0.0719 \\
  \bottomrule
  \end{tabular}
  \label{tab:ablation_study_pacth_size}
\end{table}

\begin{table}[t]
  \centering
  \caption{Variants of video resolution.  }
  \scalebox{0.95}{\begin{tabular}{@{}c|ccc|ccc@{}}
    \toprule
     \multirow{2}*{\bf Resolution} & \multicolumn{3}{c|}{\bf $\times$2} & \multicolumn{3}{c}{\bf $\times$4}\\
     & {\bf $\Delta$PSNR}   & {\bf $\Delta$VMAF} & {\bf $\Delta$LPIPS}        & {\bf $\Delta$PSNR}   & {\bf $\Delta$VMAF}  & {\bf $\Delta$LPIPS} \\
    \midrule
              1280 $\times$ 720   & $+$2.52 & $+$3.18 &  $-$0.0122 & $+$1.97 & $+$7.82 & $-$0.0683 \\
             1920 $\times$ 1080 & $+$2.84 & $+$3.67 &  $-$0.0145 & $+$2.11 & $+$8.45 & $-$0.0712 \\
             3840 $\times$ 2160    & $+$3.14 & $+$4.34 &  $-$0.0183 & $+$2.37 & $+$9.36 & $-$0.0905 \\
  \bottomrule
  \end{tabular}}
  \label{tab:ablation_resolution}
\end{table}

\begin{table}[t]
  \centering
   \caption{Variants of Quantization Parameter (QP).}
  \begin{tabular}{@{}c|ccc|ccc@{}}
     \toprule
    \multirow{2}*{\bf QP} & \multicolumn{3}{c|}{\bf $\times$2} & \multicolumn{3}{c}{\bf $\times$4}\\
     & {\bf $\Delta$PSNR}   & {\bf $\Delta$VMAF} & {\bf $\Delta$LPIPS}        & {\bf $\Delta$PSNR}   & {\bf $\Delta$VMAF}  & {\bf $\Delta$LPIPS} \\
     \midrule
             22 & $+$2.86 & $+$3.71 &  $-$0.0152 & $+$2.24 & $+$8.67 & $-$0.0727 \\
             27 & $+$2.84 & $+$3.67 &  $-$0.0145 & $+$2.11 & $+$8.45 & $-$0.0712 \\
             32& $+$2.27 & $+$2.91 &  $-$0.0121 & $+$2.04 & $+$7.92 & $-$0.0709 \\
              37 & $+$1.65 & $+$2.04 &  $-$0.0108 & $+$1.51 & $+$6.37 & $-$0.0578 \\
   \bottomrule
  \end{tabular}
  \label{tab:variantes_of_quanntization_parameter}
\end{table}

\begin{table*}[!h]
% \vspace{-0.7em}
   \linespread{0.9}
   % \scriptsize

  % \vspace{-0.9em}
  \centering
 \caption{Average PSNR/VMAF for LR videos using VVC codec.}
  \scalebox{0.86}{
\begin{tabular}{c|l|r|ccc|ccc|ccc|ccc|ccc}
    \toprule
        \multirow{2}*{ Scale} & \multirow{2}*{ Method} & \multirow{2}*{ Patches} & \multicolumn{3}{c|}{\bf FSRCNN} & \multicolumn{3}{c|}{\bf ESPCN} & \multicolumn{3}{c|}{\bf CARN} & \multicolumn{3}{c|}{\bf WDSR} & \multicolumn{3}{c}{\bf SwinIR} \\
      & &  & PSNR   & VMAF & LPIPS  & PSNR   & VMAF  & LPIPS & PSNR  & VMAF  & LPIPS  & PSNR    & VMAF& LPIPS & PSNR    & VMAF & LPIPS \\
    \midrule
      \multirow{5}*{$\times$ 2} & agDNN~\cite{agDNN_2017} &  0 \% & 27.86 & 79.39 & 0.1173 & 30.51 & 78.27 & 0.1089 & 34.54 & 81.15 & 0.1023 & 34.07 & 84.16 & 0.1027 & 34.87 & 85.91 & 0.0956 \\
      & NAS~\cite{NAS_2018} &  100.00 \% & 33.33 & 81.75 & 0.1011 & 31.72 & 81.88 & 0.1025 & 35.81 & 83.84 & 0.0979 & 35.10 & 88.81 & 0.0915 & 37.41 & 88.39 & 0.0774 \\
      & LiveNAS~\cite{LiveNAS_2020} &  17.09 \% & 27.90 & 79.60 & 0.1156 & 31.47 & 79.63 & 0.1050 & 34.87 & 82.02 & 0.1013 & 34.57 & 87.33 & 0.0928 & 35.14 & 86.64 & 0.0899 \\
      & EMT~\cite{EMT_2022} &  17.09 \% & 33.52 & 82.46 & 0.0992 & 32.32 & 81.97 & 0.0957 & 35.08 & 82.97 & 0.0991 & 34.71 & 88.35 & 0.0920 & 36.73 & 87.06 & 0.0806 \\
      & \textbf{EPS (ours)} &  17.09 \% & 33.74 & 82.54 & 0.0987 & 32.85 & 82.24 & 0.0932 & 35.98 & 84.14 & 0.0951 & 35.87 & 89.04 & 0.0904 & 37.63 & 89.29 & 0.0770 \\
    \midrule
      \multirow{5}*{$\times$ 4} & agDNN~\cite{agDNN_2017} &  0 \% & 26.51 & 45.79 & 0.3598 & 27.77 & 41.90 & 0.3754 & 29.28 & 58.87 & 0.3307 & 29.98 & 61.35 & 0.3452 & 29.49 & 62.11 & 0.3308 \\
      & NAS~\cite{NAS_2018} &  100.00 \% & 29.26 & 49.89 & 0.3281 & 28.66 & 48.78 & 0.3504 & 30.50 & 62.03 & 0.3020 & 30.70 & 69.55 & 0.3305 & 31.53 & 66.09 & 0.2761 \\
      & LiveNAS~\cite{LiveNAS_2020} &  24.31 \% & 28.41 & 46.34 & 0.3344 & 27.98 & 45.09 & 0.3656 & 30.03 & 61.08 & 0.3111 & 30.12 & 64.45 & 0.3431 & 29.90 & 63.35 & 0.3252 \\
      & EMT~\cite{EMT_2022} &  24.31 \% & 29.36 & 49.82 & 0.3326 & 28.22 & 48.24 & 0.3550 & 30.89 & 63.66 & 0.2987 & 30.26 & 67.28 & 0.3325 & 30.99 & 65.15 & 0.2973 \\
      & \textbf{EPS (ours)} &  24.31 \% & 29.59 & 51.17 & 0.3227 & 28.71 & 50.95 & 0.3492 & 30.76 & 65.51 & 0.2952 & 30.62 & 69.48 & 0.3309 & 31.46 & 65.82 & 0.2794 \\
  \bottomrule
  \end{tabular}
}
  \label{tab:vvc}
\end{table*}

\noindent $\rhd$ \textbf{Impact of Target Video Resolution.} To evaluate the scalability and practical applicability of our EPS framework across diverse streaming scenarios, we conducted additional experiments targeting three distinct HR specifications: 720p, 1080p, and 4K. As demonstrated in Table~\ref{tab:ablation_resolution}, our method maintains consistent effectiveness across all scales. While the performance gains at 720p are slightly more modest, they remain strictly positive and substantial. Notably, as the target HR scales up to 4K, our EPS framework exhibits even greater performance gains. Consequently, our method proves highly scalable and particularly advantageous for modern ultra-high-definition video delivery.

\noindent \textbf{ $\rhd$ Impact of Quantization Parameter.} To investigate the impact of input LR video qualities on the training effectiveness of the SR model, we conduct experiments to handle the compressed video datasets with QP = \{22, 27, 32, 37\}. For each QP, we train all models to evaluate their SR performance and report the average training gain compared to the pre-trained model. The results in Table~\ref{tab:variantes_of_quanntization_parameter} show that as the QP increases, the quality of the SR video gradually decreases since the LR video introduces more distortion. However, our method can still achieve promising training improvement at high QP (i.e., QP=37).

\noindent \textbf{ $\rhd$ Impact of Video Codec. } In this part, we provide the experimental results for LR videos encoded with the VVC codec. Specifically, we encode the LR video using the VVenC encoder with the medium preset and a QP of 27. As shown in Table~\ref{tab:vvc}, our method achieves the highest PSNR and VMAF compared to other approaches across all SR backbones. Our experiments with VVC achieved even greater improvements than those with HEVC because the quality of the LR video was higher. Our experiments with HEVC also show that higher-quality LR videos, such as those with QP 22 in Table~\ref{tab:variantes_of_quanntization_parameter}, result in greater improvements.

\section{Conclusion}
In this paper, we propose an efficient patch sampling method for leveraging the overfitting property of DNNs in content-aware training for video SR models. To reduce computational costs while maintaining the overfitting quality, we sample the most informative patches from video frames to accelerate training. To achieve this, we initially partition frames into non-overlapping patches and assess texture and motion complexity using two DCT-based metrics: $SF$ (spatial feature) and $TF$ (temporal feature). Subsequently, for each frame, we group $SF$ and $TF$ values into $N$ clusters and select patches belonging to the $N^{th}$ cluster in both $SF$ and $TF$.
Our approach achieves comparable or even superior SR quality performance compared to models trained with all data, thus significantly reducing the training input.  
Our approach undergoes extensive experimentation across diverse video content involving 120 video sequences. We assess its effectiveness and generalizability by employing five SR architectures: FSRCNN, ESPCN, CARN, WDSR, and SwinIR. Our patch sampling approach is observed to reduce the number of training patches by 75.00\% to 91.69\%, depending on the resolution and the input number of clusters (\(N\)). Compared to the state-of-the-art, it selects more informative patches for training while achieving a speedup of 2.1$\times$ to 82.1$\times$ in patch sampling. Furthermore, our approach demonstrates efficiency across several ablation studies, which examine the effects of patch features, clustering strategy, number of clusters, training epochs, patch sizes, quantization parameters, and video codecs.

%%%%%%%%% REFERENCES
\balance
\bibliographystyle{IEEEtran}

\begin{IEEEbiography}[{\includegraphics[width=0.9in,height=1in,keepaspectratio]{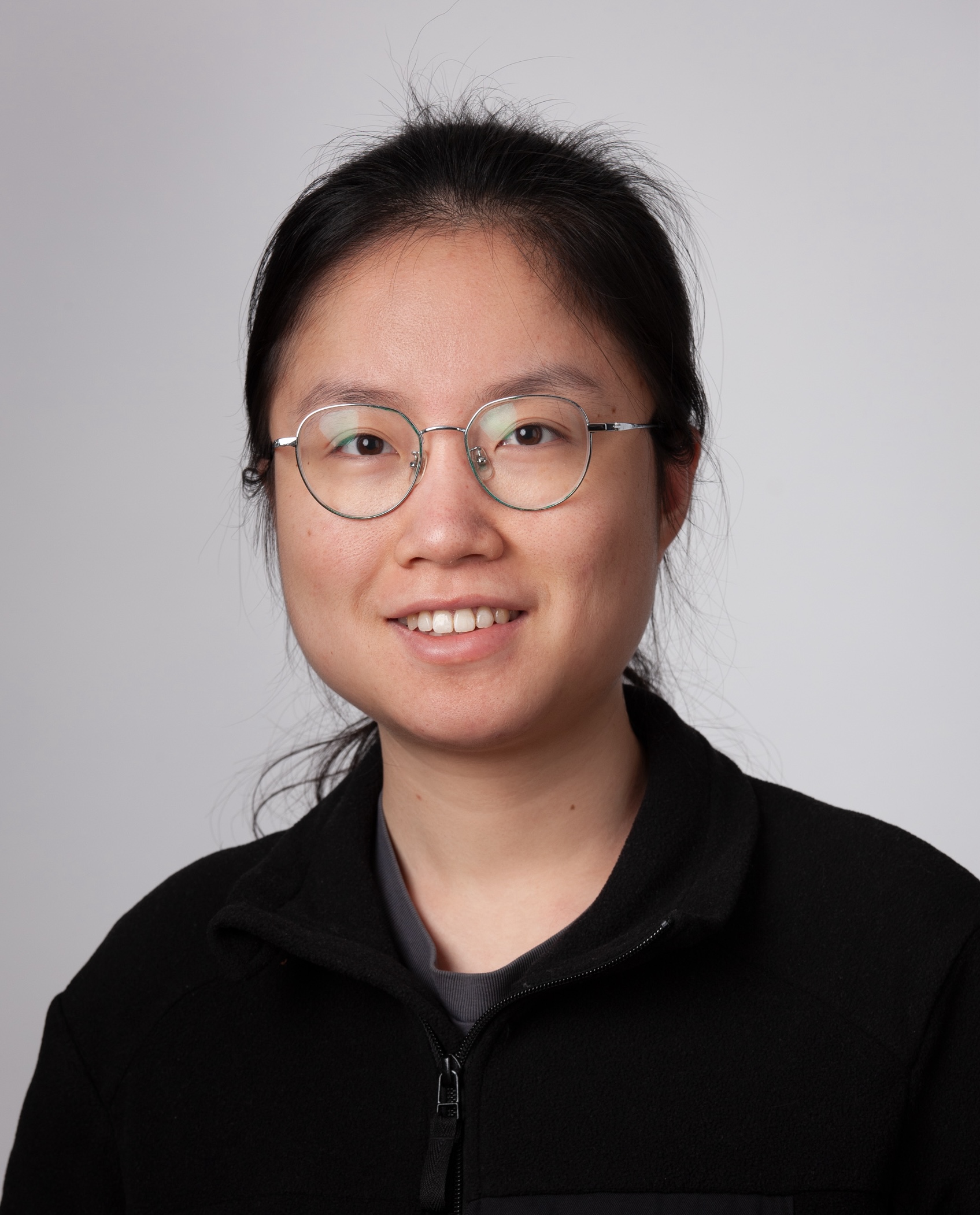}}]{Yiying Wei} is currently pursuing the Ph.D. degree at the Institute of Information Technology (ITEC), University of Klagenfurt, Austria. She received the M.S. degree in Electronic and Digital Technology from the University of Nantes, France, and the B.S. degree in Information Engineering from the South China University of Technology. Her research interests include video compression and video streaming using deep learning.
\end{IEEEbiography}

\begin{IEEEbiography}[{\includegraphics[width=0.9in,height=1in,keepaspectratio]{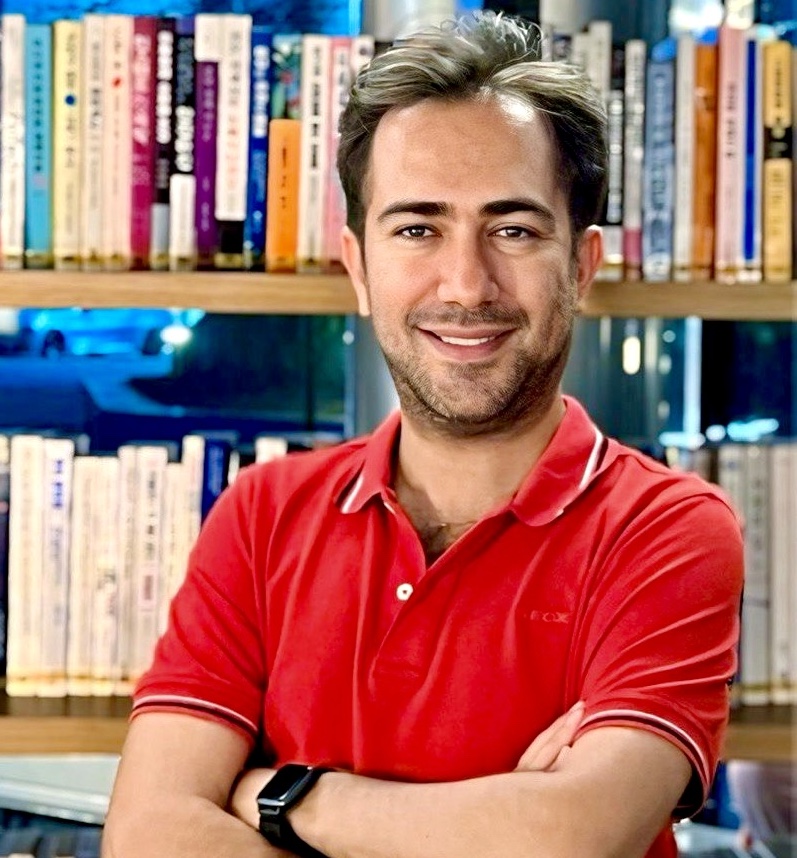}}]{Hadi Amirpour} is a tenure-track assistant professor at the Institute of Information Technology (ITEC), University of Klagenfurt, Austria.  His research interests include video streaming, image and video compression, quality of experience, emerging 3D imaging technologies, and medical image analysis.    He has received the Best Paper Awards from PCS 2024 and NAB 2025, and he serves as an Associate Editor of IEEE TCSVT and ACM TOMM.  Furthermore, he has played an active role in organizing conferences, special sessions, workshops, and tutorials at leading international events, including VCIP 2025, IEEE ICME 2025, IEEE QoMEX 2025, ACM MM 2025, VQEG 2024, IEEE ICME 2023, ACM MM 2022, IEEE EUVIP 2022, and ACM MobiSys 2022. Further information at \url{https://hadiamirpour.github.io}.
\end{IEEEbiography}

\begin{IEEEbiography}[{\includegraphics[width=0.9in,height=1in,keepaspectratio]{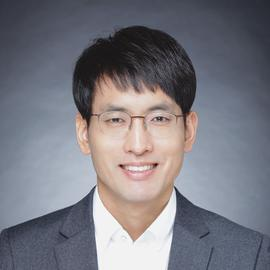}}] {Jong Hwan Ko} received the dual B.S. degree in computer science and engineering and mechanical and aerospace engineering, the M.S. degree in electrical engineering and computer science from Seoul National University, Seoul, South Korea, and the Ph.D. degree from the School of Electrical and Computer Engineering, Georgia Tech, in 2018. During his seven years research experience at the Agency for Defense Development (ADD), South Korea, he conducted advanced research on the design and performance analysis of military wireless sensor networks. He joined Sungkyunkwan University (SKKU), South Korea, as an Assistant Professor. His research interests include the design of low-power image sensor systems and deep learning accelerators for efficient image/audio processing. He has received the Best Paper Award from the International Symposium on Low Power Electronics and Design (ISLPED) in 2016.
\end{IEEEbiography}

\begin{IEEEbiography}[{\includegraphics[width=0.9in,height=1in,keepaspectratio]{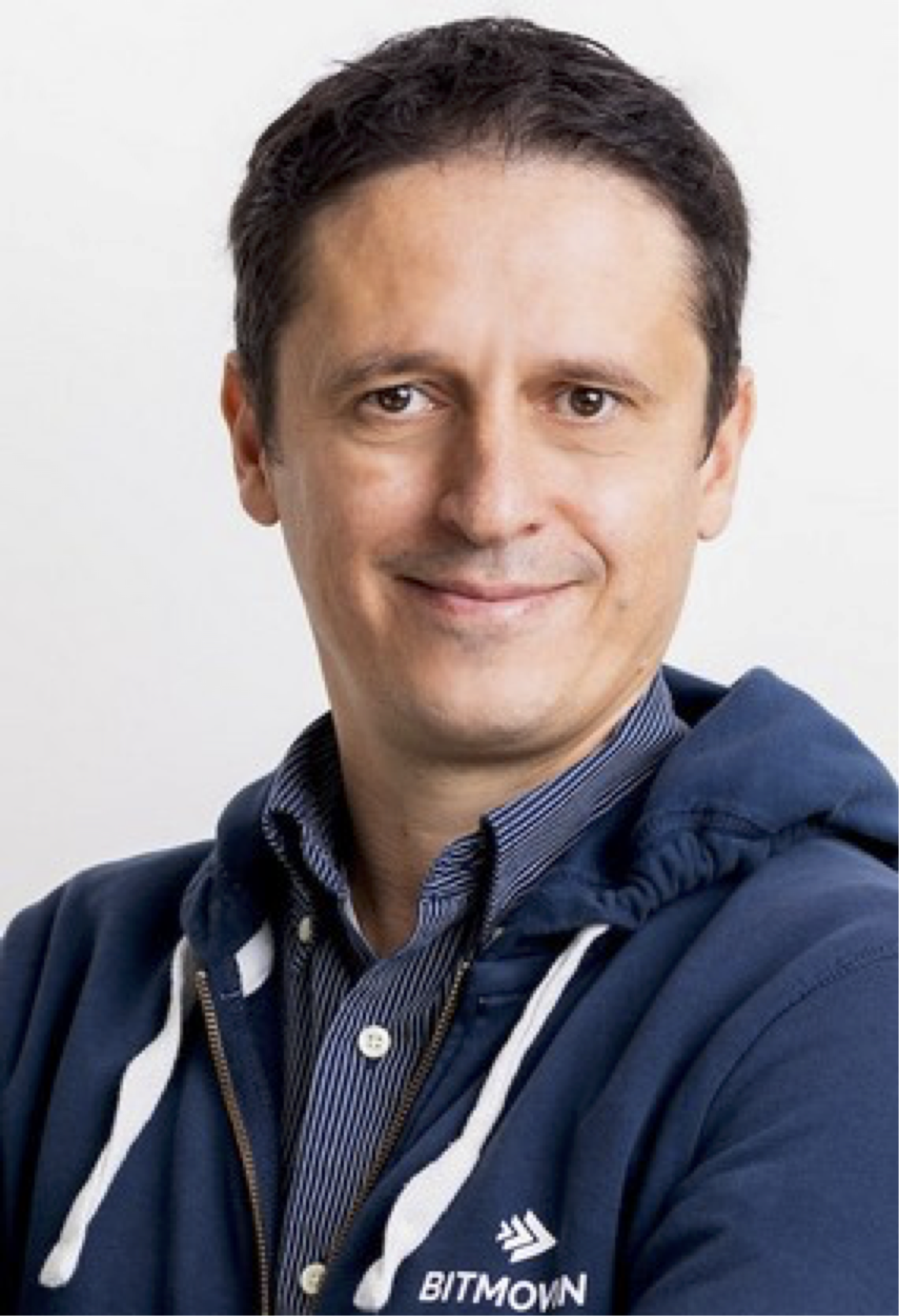}}]{Christian Timmerer} (M'08-SM'16) is a full professor of computer science at the Institute of Information Technology (ITEC) and is the director of the Christian Doppler (CD) Laboratory ATHENA (\url{https://athena.itec.aau.at/}). His research interests include immersive multimedia communication, streaming, adaptation, and quality of experience, where he co-authored seven patents and more than 300 articles. He was the general chair of WIAMIS 2008, QoMEX 2013, MMSys 2016, and PV 2018 and has participated in several EC-funded projects, notably DANAE, ENTHRONE, P2P-Next, ALICANTE, SocialSensor, COST IC1003 QUALINET, ICoSOLE, and SPIRIT. He also participated in ISO/MPEG work for several years, notably in the area of MPEG-21, MPEG-M, MPEG-V, and MPEG-DASH where he also served as standard editor. In 2013 he cofounded Bitmovin (\url{http://www.bitmovin.com/}) to provide professional services around MPEG-DASH where he holds the position of the Chief Innovation Officer (CIO) –- Head of Research and Standardization. 
Further information at \url{http://timmerer.com}. 
\end{IEEEbiography}

\end{document}